%% file: arxiv_paper.tex
\documentclass{article}

\usepackage{amsmath,amssymb}
\usepackage{color}
\usepackage{arxiv}
\input{content/preamble.tex}
\setboolean{preprint}{true}

\title{\getTitle}

\date{}

\author{ \href{https://orcid.org/0000-0002-5553-1007}{\includegraphics[scale=0.06]{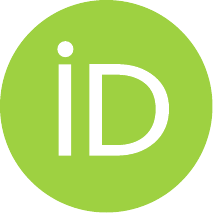}\hspace{1mm}Philipp Reiser}\thanks{Corresponding author: \href{mailto:philipp-luca.reiser@simtech.uni-stuttgart.de}{philipp-luca.reiser@simtech.uni-stuttgart.de}.} \\
	Cluster of Excellence SimTech\\
	University of Stuttgart\\
	Germany
    \And
	\href{https://orcid.org/0000-0001-5765-8995}{\includegraphics[scale=0.06]{orcid.pdf}\hspace{1mm}Paul-Christian Bürkner} \\
	Department of Statistics\\
	TU Dortmund University\\
	Germany
 	\And
	\href{https://orcid.org/0000-0003-2901-1603}{\includegraphics[scale=0.06]{orcid.pdf}\hspace{1mm}Anneli Guthke} \\
	Cluster of Excellence SimTech\\
	University of Stuttgart\\
	Germany
}

\renewcommand{\shorttitle}{\getShortTitle}

\hypersetup{
pdftitle={\getTitle},
pdfauthor={Philipp Reiser, Paul-Christian Bürkner, Anneli Guthke},
pdfkeywords={\getFirstKeyword, \getSecondKeyword, \getThirdKeyword, \getFourthKeyword},
}

\begin{document}
\maketitle

\begin{abstract}
    \input{content/abstract.tex}
\end{abstract}

\keywords{\getFirstKeyword \and \getSecondKeyword \and \getThirdKeyword \and \getFourthKeyword}

\input{content/main}

\section*{Acknowledgments}
\input{content/acknowledgments.tex}

\bibliographystyle{unsrtnat}
\bibliography{references}  
\clearpage
\appendix
\input{content/appendix.tex}

\end{document}

%% file: content/preamble.tex
\usepackage[utf8]{inputenc} 
\usepackage[T1]{fontenc}    
\usepackage{hyperref}       
\usepackage{url}            
\usepackage{booktabs}       
\usepackage{amsfonts}       
\usepackage{nicefrac}       
\usepackage{microtype}      
\usepackage{cleveref}       
\usepackage{lipsum}         
\usepackage{graphicx}
\usepackage{natbib}
\usepackage{doi}
\usepackage{arydshln}    
\usepackage{enumitem}
\usepackage{algorithm}
\usepackage{setspace}
\usepackage[noend]{algpseudocode}
\usepackage{ifthen}
\usepackage{placeins}

\newcommand{\nE}{\mathbb{E}}

\newcommand{\nN}{\mathbb{N}}

\newcommand{\nR}{\mathbb{R}}

\newcommand{\cD}{\mathcal{D}}

\newcommand{\cG}{\mathcal{G}}

\newcommand{\cM}{\mathcal{M}}

\newcommand{\figref}[1]{Fig.~\ref{#1}}
\newcommand{\secref}[1]{Section~\ref{#1}}
\newcommand{\eqnref}[1]{Eq.~\eqref{#1}}

\newcommand{\algoref}[1]{Algorithm~\ref{#1}}

\makeatletter
\DeclareRobustCommand\onedot{\futurelet\@let@token\@onedot}
\def\@onedot{\ifx\@let@token.\else.\null\fi\xspace}

\makeatother

\definecolor{darkred}{rgb}{0.8,0,0}
\definecolor{darkgreen}{rgb}{0,0.5,0}
\definecolor{darkblue}{rgb}{0,0,0.7}
\definecolor{darkpurple}{rgb}{0.4,0,0.6}
\definecolor{lightgray}{rgb}{0.92,0.92,0.92}
\definecolor{lightpink}{rgb}{1.00,0.90,0.90}

\newcommand*{\getTitle}{Bayesian Surrogate Training on Multiple Data Sources: A Hybrid Modeling Strategy}
\newcommand*{\getShortTitle}{Bayesian Surrogate Training on Multiple Data Sources}

\newcommand*{\getFirstKeyword}{Surrogate Modeling}
\newcommand*{\getSecondKeyword}{Hybrid Modeling}
\newcommand*{\getThirdKeyword}{Uncertainty Quantification}
\newcommand*{\getFourthKeyword}{Machine Learning}

\newboolean{preprint}

%% file: content/abstract.tex
Surrogate models are often used as computationally efficient approximations to complex simulation models, enabling tasks such as solving inverse problems, sensitivity analysis, and probabilistic forward predictions, which would otherwise be computationally infeasible. During training, surrogate parameters are fitted such that the surrogate reproduces the simulation model's outputs as closely as possible. However, the simulation model itself is merely a simplification of the real-world system, often missing relevant processes or suffering from misspecifications e.g., in inputs or boundary conditions. Hints about these might be captured in real-world measurement data, and yet, we typically ignore those hints during surrogate building. In this paper, we propose two novel probabilistic approaches to integrate simulation data and real-world measurement data during surrogate training. The first method trains separate surrogate models for each data source and combines their predictive distributions, while the second incorporates both data sources by training a single surrogate. Both hybrid modeling approaches employ a novel weighting strategy for combining heterogeneous data sources during surrogate training, which operates independently of the chosen surrogate family. We show the conceptual differences and benefits of the two approaches through both synthetic and real-world case studies. The results demonstrate the potential of these methods to improve predictive accuracy, predictive coverage, and to diagnose problems in the underlying simulation model. These insights can improve system understanding and future model development. 

%% file: content/main.tex
\section{Introduction}
Surrogate models are widely used as computationally efficient approximations to complex simulation models. Their applications span diverse fields, including hydrology \citep{mohammadiBayesianSelectionHydromorphodynamic2018, tarakanov2019regressionbasedsparsepolynomial}, and systems biology \cite{renardy2018parameteruncertaintyquantification, alden2020usingemulationengineer}, where simulation models are critical for understanding complex systems and making predictions. Examples of surrogate models include Polynomial Chaos Expansions (PCE) \citep{wiener1938homogeneouschaos, sudret2008globalsensitivityanalysis, oladyshkin2012datadrivenuncertaintyquantification, burknerFullyBayesianSparse2023a}, Gaussian Processes (GPs) \citep{kennedyBayesianCalibrationComputer2001, rasmussen2008gaussianprocessesmachine}, and Neural Networks \citep{Goodfellow-et-al-2016}. Surrogates are often employed to replace expensive simulations in tasks such as uncertainty quantification or solving inverse problems \citep{li2014adaptive, reiser2025uncertaintyquantificationpropagation}, sensitivity analysis, or probabilistic forward prediction \citep{ranftl2021bayesiansurrogateanalysis}. However, a significant limitation arises when the simulation model, to which the surrogate is traditionally fitted, is itself an imperfect approximation of real-world processes. In such cases, the surrogate model inherits the inaccuracies of the simulation model, limiting its predictive capabilities.

To address this limitation, there has been growing interest in integrating real-world measurement data into surrogate modeling.  However, the inherent asymmetry between the two data sources complicates the integration process. For simulation data, all input parameters are specified, whereas some of them may be unknown for real-world measurement data. There are several existing approaches for integrating real-world data. The seminal work on Bayesian calibration of computer models \citep{kennedyBayesianCalibrationComputer2001} introduces an explicit, typically input-dependent discrepancy model to capture deviations between the simulation model and real-world data. This framework provides a principled approach to combining both data sources, but its practical implementations usually rely on GPs and simplifying assumptions to make full Bayesian calibration computationally feasible \citep{bayarriModularizationBayesianAnalysis2009}. Other approaches adaptively query the simulation model in regions of high posterior densities \cite{li2014adaptive, scheurerSurrogatebasedBayesianComparison2021} in order to solve inverse problems more efficiently, but do not train the surrogate directly with real-world data. Recently, \cite{bridgmanEnhancingPolynomialChaos2023} proposed a method to directly train polynomial surrogate models with two data sources using a transfer learning perspective but assuming homogeneous data sources with the same known inputs.
Also other lines of research tackle the same problem, utilizing approximate Gaussian filtering and smoothing \citep{Särkkä_Svensson_2023, schmidt2021probabilisticstatespace} or probabilistic differential equation solvers \citep{tronarp2019probabilisticsolutionsordinary, bosch2021calibratedadaptiveprobabilistic}. 
However, these existing methods are all limited by not directly using real-world data for surrogate training, assuming homogeneous data sources, or making unrealistic normality assumptions. 

What is more, the question of how to appropriately weight different data sources during surrogate training has received limited attention in the literature. Such weighting allows to reflect varying levels of confidence or trust in the data. Simulation data encode well-established mechanistic knowledge but may suffer from structural misspecification, while real data are more trustworthy but often scarce, noisy, or subject to unknown biases. Choosing an appropriate weighting strategy is therefore crucial, as it can significantly influence the predictions of the surrogate. 

Motivated by these challenges, we propose a general-purpose framework for training Bayesian surrogate models on multiple data sources. We introduce weighting-based methods that enable modelers to flexibly control the relative influence of simulation and real-world data during surrogate training. This framework applies to arbitrary surrogate model classes that can be implemented in a probabilistic programming language, for example using Stan \citep{standev2024stan} or PyMC \citep{oriol2023pymc}.
Specifically, we present two novel probabilistic approaches for weighted integration of simulation and real-world data during surrogate training: (a) Posterior predictive weighting, which combines the predictions of separately trained surrogates using a weighted sum of their posterior predictive distributions. (b) Weighting by power-scaling the likelihood, which trains a single surrogate by adjusting the influence of each data source directly through power-scaling the likelihoods. The first method overcomes the discussed challenges by independently training surrogates on the heterogeneous data sources. The second method addresses asymmetries in the data through a two-step procedure: First, it jointly estimates the surrogate coefficients from both data sources and subsequently refines the inference of the unknown input variables for the real-world measurement data. These approaches not only tackle existing limitations, but also enable to improve predictive capabilities in challenging out-of-distribution regimes as we show in our case studies. Further, the proposed approaches extend the suite of existing methods for hybrid modeling (i.e., merging simulation with real-world data) and offer diagnostic insights into the complementary information provided by the different data sources, and into potential structural limitations of the simulation model.

While we derive the proposed equations specifically for the two obvious data classes simulation data and real-world data, our framework is general in the sense that it could accommodate multiple, arbitrary data sources (e.g., differentiating between simulation or real-world data of different fidelity or between different data types). We are confident that this generalized Bayesian framework for surrogate training on multiple data sources will prove useful in a wide range of disciplines, as it promotes computationally efficient surrogates (of arbitrary type) from being solely a ``replacement'' to being an ingredient of a powerful diagnostic tool for hybrid modeling.

The remainder of the paper is organized as follows. In \secref{sec:methods}, we provide an overview of existing surrogate modeling techniques and present the two novel approaches for weighted surrogate training on multiple data sources. In \secref{sec:case_studies}, we illustrate the conceptual differences between our approaches and their ability to improve predictive skills. Specifically, we evaluate the approaches in two case studies, ranging from synthetic scenarios to real-world problems with varied scenarios of available measurement data. \secref{sec:conclusion} draws conclusions and provides avenues for further research.

\section{Methods}
\label{sec:methods}

\begin{figure*}[ht]
    \includegraphics[width=\textwidth]{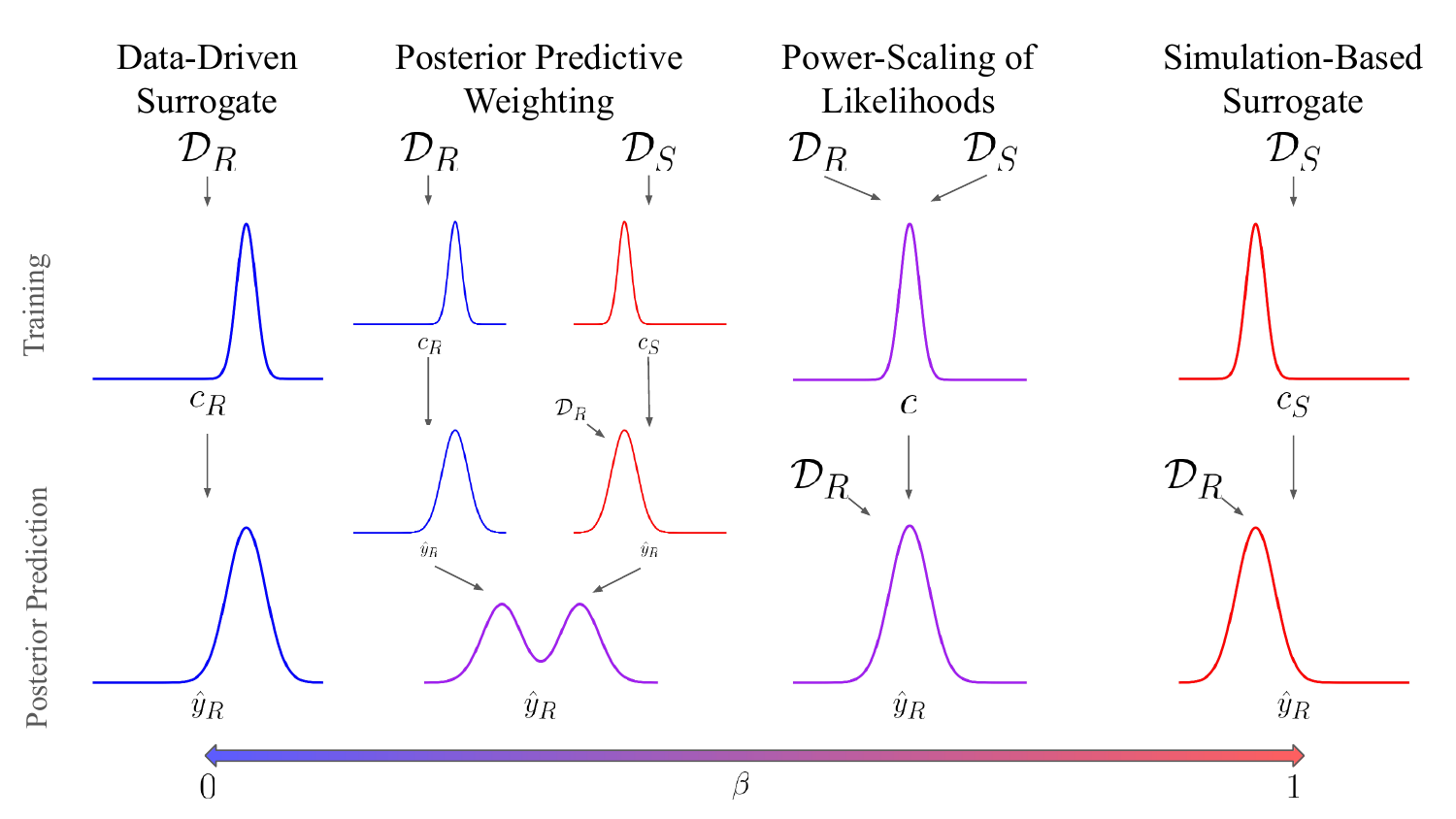}
    \caption{Schematic overview of the four discussed surrogate approaches (four columns from left to right). In the top row, we illustrate the training process of each surrogate model, indicating the data sources used: either simulation data $\cD_S$, real data $\cD_R$, or both. The posterior (solid colored line) distributions of the surrogate model coefficients are displayed, where the posterior depicts the outcome of the training process based on the used data. In the bottom row, we show the posterior predictions of the surrogate model for the output variable $\hat{y}_R$. We indicate for each approach whether an additional inference step for unknown parameters using real data $\cD_R$ is needed. Finally, the colored arrow depicts the role of the weighting factor $\beta$.
    }\label{fig:overview}
\end{figure*}

In \secref{subsec:bayesian_surrogates}, we give a brief overview of Bayesian surrogate modeling as it is commonly applied in various disciplines, and introduce a two-step procedure for surrogate training and prediction which is used throughout the paper. Next, we describe the two data sources we aim to combine: simulation data and real-world data (\secref{subsec:data_sources}). The first surrogate type, the \textit{simulation-based surrogate model} (introduced in \secref{subsec:simulation_surrogate}), is a typical surrogate focused on approximating a complex simulation model. The second surrogate type, the \textit{data-driven surrogate model}, is a purely data-driven model which is trained only using real-world data (introduced in \secref{subsec:data_driven_surrogate}). To integrate both data sources, we propose two \textit{hybrid surrogate} modeling approaches: (i) weighting of posterior predictive distributions (\secref{subsubsec:pp_weighting}) and (ii) power-scaling the likelihoods (\secref{subsubsec:power_scaling}). \figref{fig:overview} provides an illustration of these four surrogate approaches.

\subsection{Bayesian Surrogate Modeling}
\label{subsec:bayesian_surrogates}

Typically, surrogate models $\widetilde{\cM}$ aim to approximate complex simulation models $\cM$ in order to replace them in computationally expensive tasks. Common surrogate model classes include Polynomial Chaos Expansions, Gaussian Processes (GPs), and Neural Networks (NNs). Each surrogate type can be formulated such that its inputs consist of $x$ and $\omega$, where $x$ represents known inputs (e.g., spatial or temporal variables) and $\omega$ denotes unknown model parameters. In PCE, the surrogate parameters are the polynomial coefficients, and the expansion can be constructed on either both $(x, \omega)$ or solely on the unknown inputs $\omega$. For GPs, we distinguish between \textit{marginal GPs} -- which assume a Gaussian likelihood and infer hyperparameters such as the length scale and variance -- and \textit{latent GPs}, which extend this formulation to non-Gaussian likelihoods and additionally treat the latent function values as parameters. In neural network surrogates, the parameters are given by the network’s weights and biases. One such task is to solve the probabilistic inverse problem using surrogate models. This problem can be formulated as a Bayesian inference problem. \cite{reiser2025uncertaintyquantificationpropagation} propose a two-step procedure to perform surrogate-based Bayesian inference, while accounting for the additional uncertainty introduced by the surrogate itself. In the first step, a surrogate model is trained by Bayesian inference of the surrogate parameters in order to approximate the simulator. By propagating the surrogate posterior to the second step, an uncertainty-aware inference posterior of the unknown input variables can be estimated. In this paper we are interested in making predictions using Bayesian surrogate models. To use this surrogate for prediction, we can compute the posterior predictive as described in \secref{subsec:simulation_surrogate}. We are assuming that the simulation data set is fixed and given before constructing the surrogate, as opposed to approaches in iterative sampling of the expensive simulator to approximate the high-density area of the simulation-based posterior best \citep[e.g.,][]{li2014adaptive}.

\subsection{Training-Data Sources}
\label{subsec:data_sources}

First, we introduce two distinct sources of data that we aim to combine in surrogate training: simulation data from a physics-based model and real-world measurement data. Below we describe their commonalities and differences. We introduce three variables: two inputs $(x, \omega)$ and the output $y$. Their meanings and availability differ between the two data sources, which thus require asymmetric treatment in our methods.

\paragraph{Simulation Data}
To generate simulation data, a simulation model $\cM$ is used that requires two types of input variables: $x_S$ (such as spatial or temporal variables), and $\omega_S$ (variables related to properties or conditions, e.g. called ``parameters'' or ``forcings''). The subscript $S$ stands for simulation. We assume that $x_S$ is fixed (chosen by the modeler) and $\omega_S$ is uncertain and treated as a random variable. Both input variables $x_S$ and $\omega_S$ are fully specified at simulation time (i.e., given or sampled randomly from the prior distribution). Given an input pair $(x_S, \omega_S)$, where both $x_S$ and $\omega_S$ could be scalars or vectors, the simulation model $\cM$ produces an output $y_S$: 
\begin{align}
    y_S = \cM(x_S, \omega_S).
\end{align}
To form a simulation data set $\cD_S$, we repeat the process for $N_S$ different inputs:
\begin{align}
    \cD_S = \{x_S^{(i)}, \omega_S^{(i)}, y_S^{(i)}\}_{i=1}^{N_S}.
\end{align}

\paragraph{Real-World Data}
The real-world measurement data are assumed to stem from an unknown real-world data-generating process $\cG$, with fixed inputs $(x_R, \omega_R)$ and measurement noise parameter $\sigma_\epsilon$:
\begin{align}
    y_R \sim \cG(x_R, \omega_R, \sigma_\epsilon).
\end{align}
For synthetic setups, we additionally consider a synthetic truth $\mu_R = \cG(x_R, \omega_R, \sigma_\epsilon=0)$, which corresponds to the noise-free data-generating process (the true but practically unobservable state of the real system). 
In contrast to the simulation data, for real data only the input $x_R$ is known while the true input $\omega_R$ is assumed to be unknown and practically unobservable. Hence, a real-world data point is formed only by $(x_R, y_R)$. Further, $\omega_R$ could possibly live in a different space than $\omega_S$ and for some data-generating processes, $\omega_R$ might not exist at all (i.e., in such cases, $\omega_S$ would represent an effective model parameter that does not have a physical interpretation in the real-world system). 

We assume to have $N_R$ independent observations together forming the real-world dataset:
\begin{align}
    \cD_R = \{ x_R^{(j)}, y_R^{(j)} \}_{j=1}^{N_R}.
\end{align}

\subsection{Simulation-based Surrogate Model}
\label{subsec:simulation_surrogate}
A typical use case of surrogate models is to approximate complicated simulation models that describe specific physical phenomena. In the following, we term them \emph{simulation-based surrogates}. To utilize this surrogate for probabilistic predictions, a two-step procedure is applied which consists of a surrogate training step using simulation data and a surrogate inference step of unknown parameters using measurement data (for details see \cite{reiser2025uncertaintyquantificationpropagation}).

The surrogate model $\widetilde{\cM}_S$ takes the pair $(x_S, \omega_S)$ as input. Denoting its learnable parameters as $c_S$ (e.g., polynomial coefficients for a polynomial surrogate), we model surrogate predictions $\tilde{y}_S$ as:
\begin{align}
    \tilde{y}_S =\widetilde{\cM}_S(x_S, \omega_S, c_S).
\end{align}
If the surrogate is assumed to be misspecified with respect to the simulation model, the simulation output can be modeled with an approximation error term $e_S$ with parameters $\sigma_S$ (see \cite{reiser2025uncertaintyquantificationpropagation}). For example, if the error is additive, this implies:
\begin{align}
    y_S = \tilde{y}_S + e_S \quad \text{with} \quad e_S \sim p(e_S \mid \sigma_S).
\end{align}

\paragraph{Surrogate Training}
To train a fully probabilistic surrogate model, we define a likelihood $p(y_S \mid x_S, \omega_S, \sigma_S, c_S)$ of the simulation data $y_S$, given the surrogate model with its learnable parameters $\sigma_S$ and $c_S$. Adding priors for all parameters, we obtain the following Bayesian surrogate model:
\begin{align}
    y_S &\sim p(y_S \mid x_S, \omega_S, \sigma_S, c_S) \\
    (\sigma_S, c_S) &\sim p(\sigma_S, c_S)
\end{align}
leading to the joint posterior over the surrogate model coefficients $c_S$ and the approximation error parameters $\sigma_S$:

\ifthenelse{\boolean{preprint}}{
    \begin{align}
        p(\sigma_S, c_S \mid \cD_S) \propto \prod_{i=1}^{N_S} p(y_S^{(i)} \mid x_S^{(i)}, \omega_S^{(i)}, \sigma_S, c_S) \, p(\sigma_S, c_S).
    \end{align}
}{
    \begin{align}
    \begin{split}
        &p(\sigma_S, c_S \mid \cD_S) \\
        &\propto \prod_{i=1}^{N_S} p(y_S^{(i)} \mid x_S^{(i)}, \omega_S^{(i)}, \sigma_S, c_S) \, p(\sigma_S, c_S).
    \end{split}
    \end{align}
}

Since the posterior distribution is typically analytically intractable, sampling-based methods, such as Markov chain Monte Carlo (MCMC) \citep{robert_monte_2004}, can be used to approximate the posterior distribution. Examples for classes of probabilistic surrogates are Bayesian Polynomial Chaos Expansion (PCE) \citep{shaoBayesianSparsePolynomial2017, burknerFullyBayesianSparse2023a} or Gaussian Processes (GPs) \citep{rasmussen2008gaussianprocessesmachine, gramacy2020surrogates}. 

\paragraph{Posterior Predictive}
Given the posterior over surrogate parameters $p(c_S, \sigma_S \mid \cD_S)$, the next step is to infer the posterior of the unknown input parameters $\omega_R$ and error parameters $\sigma_R$, which model the discrepancy between the surrogate model and the real data, from real-world data $\cD_R$. In the following, we assume that the approximation error to the simulation is relatively low compared to the misfit with the real world data $\sigma_R$, and therefore neglect $\sigma_S$ (note that any misfit between the surrogate and the simulation together with the misfit between surrogate and real data is then treated in a lumped fashion, and 
$\sigma_R$ can be seen as the parameter of a lumped error model).
For any fixed value of surrogate coefficients $c_S$, the posterior over unknown parameters is given by:
\ifthenelse{\boolean{preprint}}{
    \begin{align}
        p(\omega_R, \sigma_R \mid \cD_R, c_S) \propto p(y_R \mid x_R, \omega_R, \sigma_R, c_S) \, p(\omega_R, \sigma_R),
    \end{align}
}{
    \begin{align}
    \begin{split}
        &p(\omega_R, \sigma_R \mid \cD_R, c_S)\\
        &\propto p(y_R \mid x_R, \omega_R, \sigma_R, c_S) \, p(\omega_R, \sigma_R),
    \end{split}
    \end{align}
}
with prior $p(\omega_R, \sigma_R)$ and likelihood $p(y_R \mid x_R, \omega_R, \sigma_R, c_S)$.
To account for the surrogate uncertainty, we use the Expected Posterior (E-Post) method \citep{reiser2025uncertaintyquantificationpropagation}, which propagates the uncertainty from the training step to the parameter inference by averaging over the posterior of the surrogate parameters $p(c_S \mid \cD_S)$ learned from the simulation data:
\ifthenelse{\boolean{preprint}}{
    \begin{align}
         p(\omega_R, \sigma_R \mid \cD_R) = \int p(\omega_R, \sigma_R \mid \cD_R, c_S) \, p(c_S \mid \cD_S) dc_S.
    \end{align}
}{
	\begin{align}
	\begin{split}
        &p(\omega_R, \sigma_R \mid \cD_R) \\
        &= \int p(\omega_R, \sigma_R \mid \cD_R, c_S) \, p(c_S \mid \cD_S) dc_S.
	\end{split}
	\end{align}
}
This integral can be easily approximated using draws $c_S^{(s)}$ from the posterior $p(c_S \mid \cD_S)$.

To predict outputs $\hat{y}_R$ for unseen real-world inputs $\hat{x}_R$, we obtain the posterior predictive distribution using the trained surrogate model and the inferred unknown parameters:
\ifthenelse{\boolean{preprint}}{
    \begin{align}
    \label{eq:posterior_pred_surrogate_real_data_inference}
        p(\hat{y}_R \mid \hat{x}_R, \cD_S, \cD_R, \widetilde{\cM}_S) = 
        \iiint p(\hat{y}_R \mid \hat{x}_R, \omega_R, \sigma_R, c_S) \, p(\omega_R, \sigma_R \mid \cD_R) \, p(c_S \mid \cD_S) \, d \omega_R  \, d \sigma_R \, d c_S.
    \end{align}
}{
	\begin{align}
	\begin{split}
        \label{eq:posterior_pred_surrogate_real_data_inference}
        &p(\hat{y}_R \mid \hat{x}_R, \cD_S, \cD_R, \widetilde{\cM}_S) \\
        &= \iiint p(\hat{y}_R \mid \hat{x}_R, \omega_R, \sigma_R, c_S) \, p(\omega_R, \sigma_R \mid \cD_R) \\
        &\quad \times p(c_S \mid \cD_S) \, d \omega_R  \, d \sigma_R \, d c_S.
	\end{split}
	\end{align}
}
The sequence of surrogate training using simulation data $\cD_S$ and inferring unknown parameters from on real data $\cD_R$ to obtain the surrogate-approximated posterior predictive distribution is illustrated in the first column of \figref{fig:overview}.

\subsection{Data-driven Surrogate Model}
\label{subsec:data_driven_surrogate}
Next, we discuss how to model the real-world data directly using a purely data-driven ``surrogate''. Mostly, such approaches are referred to as ``data-driven models'' rather than surrogates if they are not intended as substitutes of an expensive physics-based simulation model. Here, we use the term \emph{data-driven surrogate model} to unify the terminology across all discussed approaches. We will show in \secref{subsec:weighted_surrogate_training} that the purely data-driven approach is the opposite end-member of the simulation-based surrogate on an axis that describes the relative importance of the two data sources (simulation and real-world data) in surrogate training. Typical data-driven approaches span from neural networks \cite{Goodfellow-et-al-2016}, over Gaussian Processes \cite{rasmussen2008gaussianprocessesmachine} to PCE in data-driven setups \citep{torre2019datadrivenpolynomialchaos}. We focus on Bayesian data-driven models here. 

In contrast to the simulation-based surrogate model, this model uses only $x_R$ as input, because the input $\omega_R$ is unknown for real-world data. The data-driven surrogate model $\widetilde{\cM}_R$ with learnable parameters $c_R$ is defined as:
\begin{align}
    \tilde{y}_R =\widetilde{\cM}_R(x_R, c_R).
\end{align}
We model the observations $y_R$ using an approximation error term $e_R$ with parameters $\sigma_R$. For example, if the error is additive, this implies: 
\begin{align}
   y_R = \tilde{y}_R + e_R \quad \text{with} \quad e_R \sim p(e_R \mid \sigma_R).
\end{align}

\paragraph{Surrogate Training} To provide a fully probabilistic data-driven model, we form a likelihood $p(y_R \mid x_R, \sigma_R, c_R)$ of the data $y_R$, given the approximation error parameters $\sigma_R$ and coefficients $c_R$. Adding priors for all parameters results in the following Bayesian model:
\begin{align}
    y_R &\sim p(y_R \mid x_R, \sigma_R, c_R) \\
    (\sigma_R, c_R) &\sim p(\sigma_R, c_R) 
\end{align}
This implies the following posterior of the data-driven surrogate model parameters:
\ifthenelse{\boolean{preprint}}{
    \begin{align}
        p(\sigma_R, c_R \mid \cD_R) \propto \prod_{j=1}^{N_R} p(y_R^{(j)} \mid x_R^{(j)}, \sigma_R, c_R) \, p(\sigma_R, c_R).
    \end{align}
}{
	\begin{align}
	\begin{split}
        &p(\sigma_R, c_R \mid \cD_R) \\
        &\propto \prod_{j=1}^{N_R} p(y_R^{(j)} \mid x_R^{(j)}, \sigma_R, c_R) \, p(\sigma_R, c_R).
	\end{split}
	\end{align}
}

\paragraph{Posterior Predictive} The posterior predictive distribution is then given by:
\ifthenelse{\boolean{preprint}}{
    \begin{align}
    \label{eq:posterior_pred_data_driven}
        p(\hat{y}_R \mid \hat{x}_R, \cD_R, \widetilde{\cM}_R) = 
        \iint p(\hat{y}_R \mid \hat{x}_R, \sigma_R, c_R) \, p(\sigma_R, c_R \mid \cD_R) \, d \sigma_R \, d c_R.
    \end{align}
}{
	\begin{align}
	\begin{split}
        \label{eq:posterior_pred_data_driven}
        &p(\hat{y}_R \mid \hat{x}_R, \cD_R, \widetilde{\cM}_R) \\
        &= \iint p(\hat{y}_R \mid \hat{x}_R, \sigma_R, c_R) \, p(\sigma_R, c_R \mid \cD_R) \, d \sigma_R \, d c_R.
	\end{split}
	\end{align}
}
As opposed to the simulation-based surrogate, the data-driven surrogate requires no second inference step since we directly learn the surrogate parameters from the real data already in the first step.

\subsection{Weighted Surrogate Training on Multiple Data Sources}
\label{subsec:weighted_surrogate_training}
In real-world applications, combining information from both the simulation and real-world data may lead to more accurate and robust predictions. Exploring such scenarios is one of the main goals of this paper. Each data source has its own benefits and drawbacks: Simulation data may be available in bigger sizes and contains generalized domain knowledge, but often suffers from model structural errors that spoil a perfect representation of reality; real-world data can be more costly to acquire but often reveals local and specific insights about the underlying data-generating process. Another important difference is the role and knowledge of $\omega$: It is needed as input to the model function and hence known in simulation data, but unknown for real data and irrelevant when learning directly from data. This leads to an asymmetry between the two data sources. The challenge and opportunity we tackle here is to effectively integrate the two data sources into a single surrogate model. 

In this section, we present two possible solutions to combine simulation and real data probabilistically. The first approach combines the outputs of separately trained simulation-based and data-driven surrogate models by weighting their posterior predictive distributions (\secref{subsubsec:pp_weighting}). The second approach builds a single surrogate model that directly integrates both data sources by power-scaling the likelihoods of real and simulation data (\secref{subsubsec:power_scaling}).

\subsubsection{Weighting of Surrogate Posterior Predictive Distributions}
\label{subsubsec:pp_weighting}
\paragraph{Separate Training} The main ingredients of the first approach are the individual posterior predictive distributions, and hence we begin by training two models separately:

\begin{enumerate}[label=(\roman*)]
    \item a simulation-based surrogate $\widetilde{\cM}_S$ trained on simulation data and inferred input parameter posterior using real data (see \secref{subsec:simulation_surrogate}),
    \item  a data-driven surrogate model $\widetilde{\cM}_R$ trained on real data (see \secref{subsec:data_driven_surrogate}).
\end{enumerate}

\paragraph{Posterior Predictive} After training those two models, we combine their respective posterior predictive distributions for the prediction step (cf. third column of Fig. \ref{fig:overview}). Specifically, we determine the weighted sum of the posterior predictive distribution predicted by the simulation-based surrogate model (given by \eqnref{eq:posterior_pred_surrogate_real_data_inference}) and the posterior predictive distribution predicted by the data-driven surrogate model (given by \eqnref{eq:posterior_pred_data_driven}). We define:
\ifthenelse{\boolean{preprint}}{
    \begin{align}
        p_{\rm pw, \beta}(\hat{y}_R \mid \hat{x}_R, \cD_S, \cD_R) = \beta \, p(\hat{y}_R \mid \hat{x}_R, \cD_S, \cD_R, \widetilde{\cM}_S) + (1-\beta) \, p(\hat{y}_R \mid \hat{x}_R, \cD_R, \widetilde{\cM}_R),
    \end{align}
}{
	\begin{align}
	\begin{split}
        &p_{\rm pw, \beta}(\hat{y}_R \mid \hat{x}_R, \cD_S, \cD_R) \\
        &= \beta \, p(\hat{y}_R \mid \hat{x}_R, \cD_S, \cD_R, \widetilde{\cM}_S) \\
        &\quad+ (1-\beta) \, p(\hat{y}_R \mid \hat{x}_R, \cD_R, \widetilde{\cM}_R),
	\end{split}
	\end{align}
}
where the first subscript $\rm pw$ stands for \textit{predictive weighting} and the second subscript $\beta$ denotes a weighting factor $0 \leq \beta \leq 1$ that controls the relative influence of the data-driven and simulation-based surrogate model. This factor can be subjectively specified according to the modeler's choice, or formally optimized to maximize performance. Here, we will perform a sensitivity analysis over the value range of $\beta$ to demonstrate its impact on the resulting posterior predictive distribution.

\paragraph{Related Work} One method to calculate an optimal $\beta$ is posterior predictive stacking, as described by \cite{yaoUsingStackingAverage2018}. Stacking is particularly useful in an $\cM$-open setting, where the true model is not among the considered models. In such cases, stacking finds the optimal weights by optimizing scoring rules computed on out-of-distribution (OOD) data or via cross-validation. Since both surrogate models are approximate, posterior predictive stacking is well-suited to this task.

\subsubsection{Weighting by Power-Scaling of Likelihoods}
\label{subsubsec:power_scaling}

While the above approach trains the simulation-based and data-driven surrogate models separately, the \textit{power-scaling} method introduced below aims to combine both sources of information into a single surrogate model. This approach is inspired by the idea that building a ``super''-model may be preferable over doing post-hoc model weighting to allow to pool information across models, as discussed by \cite{yaoUsingStackingAverage2018}. We propose a two step power-scaling approach for weighted surrogate training. In the first step, we train the surrogate parameters using power-scaled likelihoods, controlling the influence of real and simulation data on the surrogate parameters $c$. In the second step, we refine the inference on $(\omega_R, \sigma_R)$ using only real data $\cD_R$, given the posterior of $c$ from the first step. 

\paragraph{Joint Training}
In the first step, we update the surrogate parameters with a weighted contribution of simulation and real-world data. To do this, we compute the posterior implied by power-scaling the likelihoods of the two data sources:
\ifthenelse{\boolean{preprint}}{
    \begin{align}
        p_{\rm ps}^\beta(\omega_R', \sigma', c \mid \cD_S, \cD_R) \propto \, p(y_R \mid x_R, \omega_R', \sigma', c)^{\alpha_R} \, p(y_S \mid x_S, \omega_S, \sigma', c)^{\alpha_S} \, p(\omega_R', \sigma') \, p(c).
    \end{align}
}{
	\begin{align}
	\begin{split}
        &p_{\rm ps}^\beta(\omega_R', \sigma', c \mid \cD_S, \cD_R)\\
        &\propto \, p(y_R \mid x_R, \omega_R', \sigma', c)^{\alpha_R} \, p(y_S \mid x_S, \omega_S, \sigma', c)^{\alpha_S} \\
        &\quad \times  p(\omega_R', \sigma') \, p(c).
	\end{split}
	\end{align}
}
Here, the intermediate variables $\omega_R'$ and $\sigma'$ are estimated, but are later on re-estimated using the full likelihood information of the real data (see below). We assume a single $\sigma'$ for both simulation and real data because estimating two separate error parameters $\sigma_S$ and $\sigma_R$ is prone to identification issues. Further, as explained above, $\sigma_R$ is expected to dominate in the case studies that we will present below, i.e., we assume that the surrogate model will fit the simulation model reasonably well after training, whereas larger deviations will occur when comparing posterior predictions with real measurements. Again, in that spirit, $\sigma'$ can be seen as a global parameter of a lumped error model that covers the surrogate's approximation error in reproducing the simulation data, the trained surrogate's error in predicting real data, and the unknown level of measurement noise in the real data. 

The scaling factors $\alpha_R \in [0, 1]$ and $\alpha_S \in [0, 1]$ control the influence of each data source, computed based on the weighting factor $\beta \in [0, 1]$:
\begin{align}
\begin{split}    
    \label{eq:scaling_factor}
    \alpha_S &= \left\{
    \begin{array}{@{}ll@{}}
    \frac{\beta}{1-\beta}, & \text{if}\ \beta < 0.5 \\
    1, & \text{otherwise}
    \end{array}\right. 
    \quad \text{and} \\
    \alpha_R &= \left\{
    \begin{array}{@{}ll@{}}
    1, & \text{if}\ \beta < 0.5 \\
    \frac{1-\beta}{\beta}, & \text{otherwise.}
    \end{array}\right.
\end{split}
\end{align}
This definition ensures that:
\begin{itemize}
    \item For $\beta = 0.5$, we have $\alpha_R = \alpha_S = 1$ such that both data sources contribute equally with the full likelihood information being utilized.
    \item For $\beta < 0.5$, full likelihood information of the real data is used ($\alpha_R = 1$), while the influence of the simulation data is diminished through $\alpha_S < 1$ until it is completely neglected for $\beta = 0$ implying $\alpha_S = 0$.
    \item For $\beta > 0.5$, full likelihood information of the simulation data is used ($\alpha_S = 1$), while the influence of the real data is diminished through $\alpha_R < 1$ until it is completely neglected for $\beta = 1$ implying $\alpha_R = 0$.
\end{itemize}

\paragraph{Posterior Predictive} The second step refines the posterior distribution of the real-world input and noise parameters $(\omega_R, \sigma_R)$ using only real data $\cD_R$. While the first step enabled a power-scaled posterior over $c$, the posterior over $(\omega_R', \sigma')$ was indirectly influenced by simulation data as $\sigma'$ was shared across both sources. For optimal predictive performance, the posterior of $(\omega_R, \sigma_R)$ should be based solely on real data $\cD_R$ using its full likelihood information. Therefore, we compute the posterior of input and noise parameters as in the inference step of the simulation-based surrogate model (see \secref{subsec:simulation_surrogate}):
\ifthenelse{\boolean{preprint}}{
    \begin{align}
        p_{\rm ps}^\beta(\omega_R, \sigma_R \mid \cD_R) = \int p(\omega_R, \sigma_R \mid \cD_R, c) \, p_{\rm ps}^\beta(c \mid \cD_S, \cD_R) dc.
    \end{align}
}{
	\begin{align}
	\begin{split}
        &p_{\rm ps}^\beta(\omega_R, \sigma_R \mid \cD_R) \\
        &= \int p(\omega_R, \sigma_R \mid \cD_R, c) \, p_{\rm ps}^\beta(c \mid \cD_S, \cD_R) dc.
	\end{split}
	\end{align}
}
This ensures that real-world parameters are inferred based on real-world data only, while leveraging the surrogate parameters trained with both data sources. Although the real data are used in both steps, each variable is only informed by the full likelihood information of the real data at most once. This avoids any ``double dipping'' into $\cD_R$. Specifically, in the second step, the parameters $(\omega_R, \sigma_R)$ are updated from their original prior distribution $p(\omega_R, \sigma_R)$, not from the posterior obtained in the first step $p_{\rm ps}^\beta(\omega_R', \sigma' \mid \cD_S, \cD_R)$.

The posterior predictive distribution is then computed based on the marginalized posterior from the joint training as well as the real data inference posterior from the second step:
\ifthenelse{\boolean{preprint}}{
    \begin{align}
        p_{\rm ps}^\beta(\hat{y}_R \mid \hat{x}_R, \cD_S, \cD_R) =  \iiint p(\hat{y}_R \mid \hat{x}_R, \omega_R, \sigma_R, c) \,  p_{\rm ps}^\beta(\omega_R, \sigma_R \mid \cD_R) \, p_{\rm ps}^\beta(c \mid \cD_S, \cD_R) \, d \omega_R\, d \sigma_R \, d c.
    \end{align}
}{
	\begin{align}
	\begin{split}
        &p_{\rm ps}^\beta(\hat{y}_R \mid \hat{x}_R, \cD_S, \cD_R) \\
        &=  \iiint p(\hat{y}_R \mid \hat{x}_R, \omega_R, \sigma_R, c) \,  p_{\rm ps}^\beta(\omega_R, \sigma_R \mid \cD_R) \\
        &\quad \times p_{\rm ps}^\beta(c \mid \cD_S, \cD_R) \, d \omega_R\, d \sigma_R \, d c.
	\end{split}
	\end{align}
}

\paragraph{MCMC Implementation} 
Practically, we implement the power-scaling method using a two-step MCMC procedure, as described in \algoref{algo:mcmc_power_scaling}. First, after calculating the scaling factors $\alpha_S$ and $\alpha_R$ via \eqnref{eq:scaling_factor}, we draw $S_1$ samples $(\omega_R'^{(s)}, \sigma'^{(s)}, c^{(s)})$ from the power-scaled posterior $p_{\rm ps}^\beta(\omega_R', \sigma', c \mid \cD_S, \cD_R)$. In the second step, for each $c^{(s)}$, we perform $S_2$ MCMC iterations to draw $(\omega_R^{(s, r)}, \sigma_R^{(s, r)}) \sim p_{\rm ps}^\beta(\omega_R, \sigma_R \mid \cD_R, c^{(s)})$. Each MCMC run is initialized with the corresponding sample $(\omega_R'^{(s)}, \sigma'^{(s)})$ from the first step. To reduce memory consumption, one can retain only the last sample of each MCMC run, while the rest is used to check convergence. This procedure produces posterior draws $(\omega_R^{(s)}, \sigma_R^{(s)}, c^{(s)})$, which follow the distribution $p_{\rm ps}^\beta(\omega_R, \sigma_R \mid \cD_R) \, p_{\rm ps}^\beta(c \mid \cD_S, \cD_R)$.

\begin{algorithm*}
\setstretch{1.35}
\caption{Two step MCMC for power-scaling the posterior.}\label{algo:mcmc_power_scaling}
\begin{algorithmic}
\State{\textbf{Input}: weight $\beta \in [0, 1]$, Data $(\cD_S, \cD_R)$, likelihoods $p(y_R \mid x_R, \omega_R, c, \sigma)$ and $p(y_S \mid x_R, \omega_S, c, \sigma)$; prior $p(\omega_R, \sigma_R) \, p(c)$, number of samples $S_1$ and $S_2$.}
\State{\textbf{Output}: Draws $\{(\omega_R^{(s)}, \sigma_R^{(s)}, c^{(s)})\}$ distributed according to the power-scaled posterior.}
\State{Calculate $\alpha_S$ and $\alpha_R$ via \eqnref{eq:scaling_factor}}

\For{$s = 1, \dots, S_1$}
    \State Sample $(\omega_R'^{(s)}, \sigma'^{(s)}, c^{(s)}) \sim p_{\rm ps}^\beta(\omega_R', \sigma', c \mid \cD_S, \cD_R)$
    \State Check convergence of joint training
\EndFor
\For{$s = 1, \dots, S_1$}
    \For{$r=1, \dots, S_2$}
        \State Sample $(\omega_R^{(s, r)}, \sigma_R^{(s, r)}) \sim p(\omega_R, \sigma_R \mid \cD_R, c^{(s)})$
        \State Check convergence of inference
    \EndFor
    Let $\omega_R^{(s)} = \omega_R^{(s, S_2)}$, $\sigma_R^{(s)} = \sigma_R^{(s, S_2)}$ (final state)
\EndFor
\State \textbf{return} $\{(\omega_R^{(s)}, \sigma_R^{(s)}, c^{(s)})\}$
\end{algorithmic}
\end{algorithm*}

\paragraph{Related Work}
Power-scaling in the context of posterior distributions has been explored in various contexts and in the following we will highlight a few of them. \cite{royall2003interpretingstatisticalevidence, agostinelliWeightedStrategyHandle2013, bissiri2016generalframeworkupdating} investigated the robustness of models under power-scaled likelihoods, while \cite{grunwald2017inconsistencybayesianinference, holmes2017assigningvaluepower} explored the benefits of power-scaling likelihoods for Bayesian linear models when model misspecification is present. Recently, \citep{mclatchie2024predictiveperformancepower} investigated the predictive performance of power-scaled posteriors. Additionally, power-scaling has been used for prior diagnostics and sensitivity analysis, as discussed by \citep{kallioinenDetectingDiagnosingPrior2024}. In the specific context of transfer learning from simulation to real-world data for PCE surrogates, \cite{bridgmanEnhancingPolynomialChaos2023} explored the idea of using power-scaling. In contrast to the previously mentioned methods, our proposed method considers power-scaling of likelihoods for asymmetric data sources which is of special interest in the context of surrogate models.

Our method is also related to several existing approaches for specific values of the weighting factor $\beta$. For $\beta = 1$, the approach corresponds to the simulation-based surrogate model explained in \secref{subsec:simulation_surrogate} and introduced by \cite{reiser2025uncertaintyquantificationpropagation}. For $\beta=0$, the method results in a data-driven surrogate model trained only on real data (see \secref{subsec:data_driven_surrogate}), but with the additional estimation of the input parameter $\omega_R$ to allow a smooth transition as $\beta$ increases. For $0.5 \leq \beta \leq 1$ our approach is conceptually similar to the semi-modular inference posterior proposed by \cite{CarmonaSmi2020}, who focus specifically on parameter inference of a Bayesian model divided into modules. 
In contrast, in our method, we focus on controlling the integration of different data sources in order to improve prediction. 

\subsection{Classification of Approaches in the Framework for Weighted Surrogate Training on Multiple Data Sources}
In the bottom of \figref{fig:overview}, we illustrate the role of the weighting factor $\beta$ in the two approaches for weighted surrogate training on multiple data sources:  posterior predictive weighting (\secref{subsubsec:pp_weighting}) and power-scaling (\secref{subsubsec:power_scaling})). This weighting factor $\beta$ controls the influence of the two data sources, real-world data $\cD_R$ and simulation data $\cD_S$ on the training process of these hybrid surrogates.

At the extremes of $\beta$, both approaches result in the same surrogates. For $\beta = 0$, the surrogate model is trained exclusively using real-world data $\cD_R$, resulting in the data-driven surrogate model (\secref{subsec:data_driven_surrogate}). In contrast, for $\beta = 1$, the surrogate is trained solely using simulation data $\cD_S$, leading to the simulation-based surrogate model (\secref{subsec:simulation_surrogate}). Hence, the framework for weighted surrogate training on multiple data sources in fact hosts all of the discussed surrogate approaches. This is why we will only speak of the two different weighted approaches in the following, instead of four possible surrogate training approaches distinguished so far.  

For intermediate values $0 < \beta < 1$, both approaches incorporate information from both data sources, balancing their contributions according to $\beta$. However, the surrogate training procedures by which they achieve this differ fundamentally as explained in the respective sections. Hence, also the resulting predictive distributions and their coverage of unseen test data will differ.

\subsection{Evaluation Metrics}
\label{subsec:evaluation_metrics}

To quantitatively assess and compare the predictive performance of the proposed approaches for weighted surrogate training, we compute two evaluation metrics: the Expected Log Pointwise Predictive Density (ELPD) and the Root Mean Squared Error (RMSE). These metrics are applied for both comparisons within an approach (e.g., for varying $\beta$ to find its optimal value) and between the two different approaches.

We denote posterior samples obtained from any of the approaches as $\{ \omega^{(s)}, \sigma^{(s)}, c^{(s)} \}_{s=1}^S$.

To model noisy unseen data using the surrogate model, we compute the posterior predictive of the respective approach. Generally, we can sample from the \textit{posterior predictive distribution} of the surrogate via:
\begin{align}
    \tilde{y}^{(s)} \sim p(\tilde{y} \mid x, \omega^{(s)}, \sigma^{(s)}, c^{(s)}).
\end{align}

We can also compute samples from the \textit{predictive mean posterior distribution} as
\begin{align}
    \tilde{\mu}^{(s)} = \int \tilde{y} \; p(\tilde{y} \mid x, \omega^{(s)}, \sigma^{(s)}, c^{(s)}) \; d\tilde{y},
\end{align}
which models the noise-free data-generating process.

\paragraph{ELPD}
We compute the expected log pointwise predictive density (ELPD) \cite{vehtariSurveyBayesianPredictive2012, vehtariPracticalBayesianModel2017} to evaluate predictive performance on previously unseen noisy test data $ \cD^* = \{ x_i^*, y_i^* \}_{i=1}^{N^*}$, consisting of $N^*$ inputs $x_i^*$ and corresponding noisy outputs $y_i^*$:
\ifthenelse{\boolean{preprint}}{
    \begin{align}
        \text{ELPD}(\cD^*) = \frac{1}{N^*} \sum_{i=1}^{N^*} \log p(y_i^* \mid x_i^*, \cD_S, \cD_R) = \frac{1}{N^*}\sum_{i=1}^{N^*}\log\left( \frac{1}{S}\sum_{s=1}^S p(y_i^* \mid x_i^*, \omega^{(s)}, \sigma^{(s)}, c^{(s)}) \right).
    \end{align}
}{
	\begin{align}
	\begin{split}
        &\text{ELPD}(\cD^*)\\
        &= \frac{1}{N^*} \sum_{i=1}^{N^*} \log p(y_i^* \mid x_i^*, \cD_S, \cD_R) \\
        &= \frac{1}{N^*}\sum_{i=1}^{N^*}\log\left( \frac{1}{S}\sum_{s=1}^S p(y_i^* \mid x_i^*, \omega^{(s)}, \sigma^{(s)}, c^{(s)}) \right).
	\end{split}
	\end{align}
}
Unlike the standard definition of ELPD (which has relation to the effective number of parameters and other quantities from information theory \citep{vehtariPracticalBayesianModel2017}), we normalize by the number of test points $N^*$ to facilitate comparison across unequally sized test sets. ELPD is particularly useful to evaluate the full posterior predictive distribution, taking into account uncertainty from both the surrogate model parameters and the error model. This makes the metric sensitive to how well the surrogate model captures uncertainty.

\paragraph{RMSE}
For synthetic scenarios, we can compute the posterior averaged root mean squared error (RMSE) using previously unseen noise-free test data $ \cD^* = \{ x_i^*, \mu_i^* \}_{i=1}^{N^*}$, consisting of $N^*$ inputs $x_i^*$ and corresponding noise-free outputs $\mu_i^*$:
\begin{align}
    \text{RMSE}(\cD^*) = \frac{1}{N^*}\sum_{i=1}^{N^*}\sqrt{\frac{1}{S} \sum_{s=1}^S\left(\tilde{\mu}^{(s)} - \mu_i^*\right)^2}.
\end{align}
The RMSE evaluates the surrogate model's predictive mean posterior distribution against the true values $\mu^*$, but does not account for the noise component and hence ignores $\sigma$. Thus, the RMSE focuses solely on the accuracy of the mean predictions. 

\section{Case Studies}
\label{sec:case_studies}
We aim to answer the following research questions through our case studies: (i) Do the weighted approaches improve predictive performance in terms of ELPD (how well is uncertainty modeled) and RMSE (how close is the mean predictive distribution to the true function) as compared to the traditional single-data-source training? (ii) Do the weighted approaches provide insights about system behavior and/or potential model deficits that would not have arised from the traditional single-data-source training?

We address these questions first in a synthetic case study (Section \ref{subsec:case_study_1}) and then in a real-world example of epidemiological modeling (Section \ref{subsec:case_study_2}), by evaluating the metrics as defined in \secref{subsec:evaluation_metrics} and by visualizing the posterior predictive distributions as derived in Sections \ref{subsubsec:pp_weighting} and \ref{subsubsec:power_scaling}. By varying the weighting factor $\beta$, we cover the standard approaches (simulation-based surrogate and data-driven surrogate) as well as the weighted approaches that integrate the two data sources simulation data and real-world data. All code can be found on GitHub\footnote{\url{https://github.com/philippreiser/multi-data-source-bayesian-surrogate-paper}}.

\subsection{Case Study 1}
\label{subsec:case_study_1}
The goal of the first case study is to assess our weighted approaches to surrogate modeling in a synthetic, controllable example. We construct a setup where the simulation model captures the global trend of the system behavior correctly, while the real-world data-generating process has an additional periodic effect that is not captured by the simulation model. This reflects a typical situation where the physics-based simulation model adequately represents dominant processes, but misses specific effects on shorter temporal or spatial scales that are either not well understood or sacrificed for the sake of parsimony and computational speed. 

\subsubsection{Setup}
\begin{figure*}[ht]
    \includegraphics[width=0.7\textwidth]{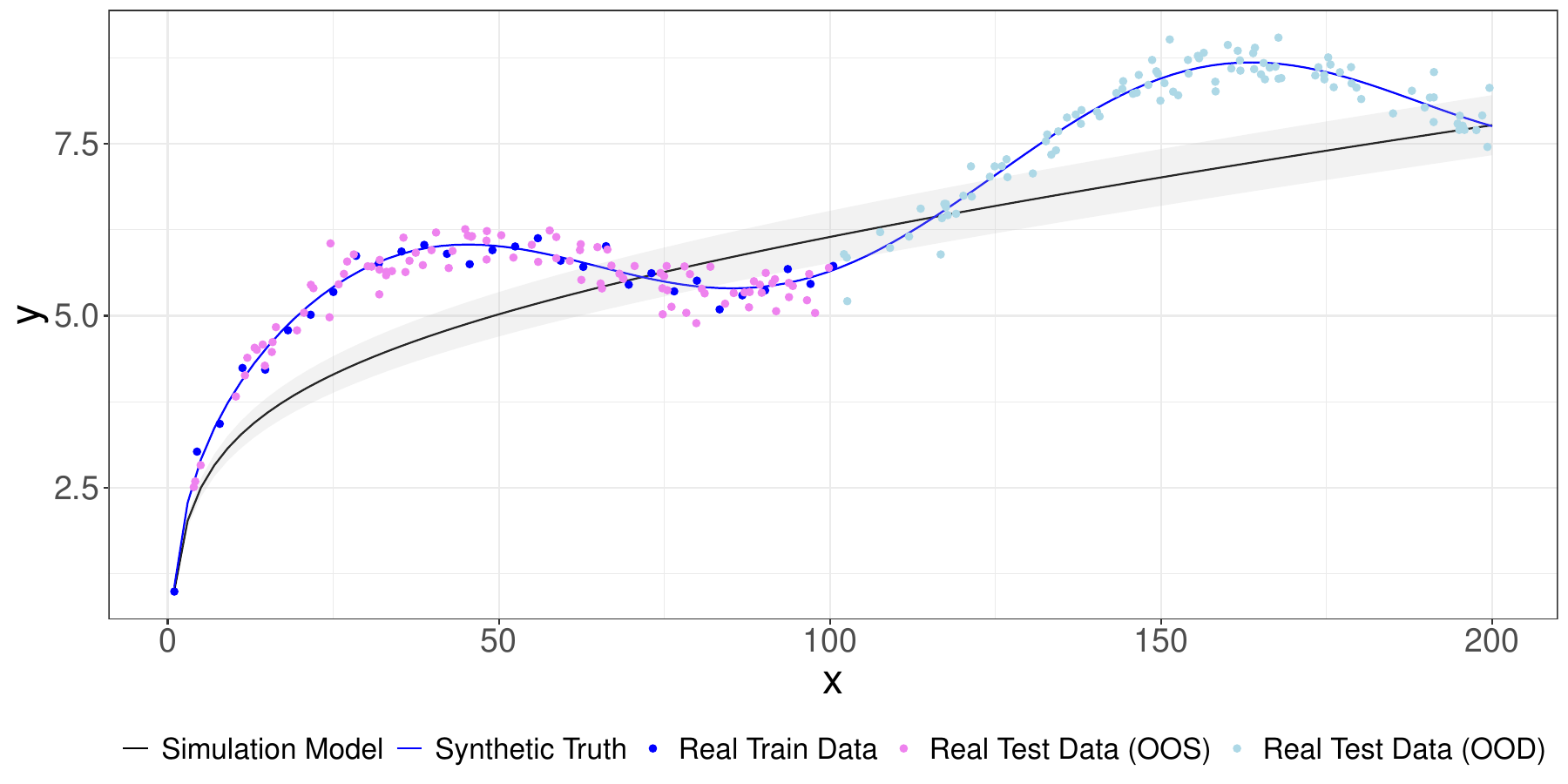}
    \centering
    \caption{Case study 1: Illustration of the simulation model and the real-world data. The black line depicts the predictive mean of the simulation model based on the prior parameter distribution $p(\omega_R)$, and the gray interval depicts the 90$\%$ credible interval of the predictive prior. The blue line depicts the synthetic truth. The dark blue data represents the (noisy) real training data, while the pink is the real OOS test data, and the light blue is the real OOD test data, respectively. }\label{fig:case_study_1_setup}
\end{figure*}
We consider the following simulation model with the known input $x_S$ and the unknown input $\omega_S$:
\begin{align}
    \cM(x_S, \omega_S) = \omega_S \log(x_S) + 0.01 x_S + 1.
\end{align}

In contrast, the real-world data-generating process is defined as:
\ifthenelse{\boolean{preprint}}{
    \begin{align}
        \cG(x_R, \omega_R=1, \sigma_\epsilon = 0.2) = \text{Normal}(\cM(x_R, \omega_R=1) + \sin(0.05 x_R), \sigma_\epsilon = 0.2)
    \end{align}
}{
	\begin{align}
	\begin{split}
        &\cG(x_R, \omega_R=1, \sigma_\epsilon = 0.2) \\
        &= \text{Normal}(\cM(x_R, \omega_R=1) \\
        &\quad + \sin(0.05 x_R), \sigma_\epsilon = 0.2)
	\end{split}
	\end{align}
}
where we assume a fixed true $\omega_R = 1$ and an additional periodic term that induces a misspecification between simulation and real system behavior. This synthetic truth is perturbed by a normally distributed measurement error with a standard deviation of $\sigma_\epsilon = 0.2$.

To generate simulation training data, we consider the first $N_S = 100$ points of the 2-dimensional Sobol sequence \citep{sobol1967distributionpointscube}, scaled to the ranges $[1, 200]$ and $[0.6, 1.4]$ for the inputs $x_S$ and $\omega_S$, respectively. These inputs are fed into the simulation model $\cM$ to generate the outputs $y_S$. Note that testing smarter or adaptive sampling schemes to build the simulation training data set for the surrogate is beyond the scope of the current study, but is recommended for future research.

To generate real-world training data, we use $N_R = 30$ evenly spaced input points in the first half of the simulation input space, i.e., $x_R \in [1, 100]$. The real-world outputs are generated by evaluating $\cG$ at the real-world inputs. This setup creates a realistic scenario with sparse and noisy real-world data. We illustrate this setup in \figref{fig:case_study_1_setup}.

As surrogate model, we use a Polynomial Chaos Expansion (PCE):
\begin{align}
    \widetilde{\cM}(x, \omega, c) = \sum_{i=0}^{d-1} c_i \psi_i(x, \omega)
\end{align}
with the vector of surrogate coefficients $c = (c_0, ..., c_{d-1})$, the total number of polynomials $d$, and multivariate Legendre polynomials $\psi_i(x, \omega)$ (see \cite{sudret2008globalsensitivityanalysis} for a detailed definition). We use two-dimensional Legendre polynomials up to a degree of 5 and follow the standard truncation scheme, leading to $d = 21$ polynomials in total. To simplify the analysis, we linearly scale the input variables and parameters to the standard scaling of the Legendre polynomials, i.e. $[-1, 1]$. 

\subsubsection{Hybrid Surrogate Modeling} 
Given a weight $\beta$, we calculate the scaling factors $(\alpha_R, \alpha_S)$ using \eqnref{eq:scaling_factor} and set the following priors and likelihoods for the power-scaling method:
\begin{align}
    y_S &\sim \text{Normal}(\widetilde{\cM}(x_S, \omega_S, c), \sigma)^{\alpha_S}\\
    y_R &\sim \text{Normal}(\widetilde{\cM}(x_R, \omega_R, c), \sigma)^{\alpha_R}\\
    \widetilde{\cM}(x, \omega, c) &= \sum_{i=0}^{d-1} c_i \psi_i(x, \omega) \\
    \omega_R &\sim \text{Normal}_{[0.6, 1.4]}(0.9, 0.05)\\
    \sigma &\sim \text{Half-Normal}(0.5)\\
    c_i &\sim \text{Normal}(0, 5).
\end{align}
Expressions such as $y \sim p(y)^{\alpha}$ denote power-scaling of the corresponding probability density function, not of the probability distribution itself. The joint power-scaled likelihood is:

\ifthenelse{\boolean{preprint}}{
\begin{align}
    p^\beta_\text{ps}(y_R, y_S \mid \cD_R, \cD_S, c)
    = \text{Normal}(\widetilde{\cM}(x_S, \omega_S, c), \sigma)^{\alpha_S} \, \text{Normal}(\widetilde{\cM}(x_R, \omega_R, c), \sigma)^{\alpha_R}.
\end{align}
}{
\begin{align}
\begin{split}
    &p^\beta_\text{ps}(y_R, y_S \mid \cD_R, \cD_S, c)\\ 
    &= \text{Normal}(\widetilde{\cM}(x_S, \omega_S, c), \sigma)^{\alpha_S} \\
    &\quad \times \text{Normal}(\widetilde{\cM}(x_R, \omega_R, c), \sigma)^{\alpha_R}.
\end{split}
\end{align}
}

For the posterior predictive weighting method, we fit a simulation-based and data-driven model using the same likelihoods and priors as specified in the probabilistic model above. In the case of the simulation-based surrogate, we neglect the surrogate approximation error $\sigma_S$, which is fitted in the first training stage, during the subsequent inference step. This decision (see \secref{subsec:simulation_surrogate}) is justified by the observation that $\sigma_R$ is an order of magnitude larger than $\sigma_S$, making the latter negligible in comparison.

We sample from the respective implied posteriors using MCMC, specifically using the adaptive Hamiltonian Monte Carlo sampler as implemented in Stan \cite{carpenter2017stan, standev2024stan}. For the first step MCMC, we use 4 chains, 1,000 warmup iterations and 250 sampling iterations, and for each draw, we run the second step MCMC with the same setting. We perform standard MCMC convergence checks, such as $\widehat{R}$ \citep{vehtari2021ranknormalizationfoldinglocalization}, applying a convergence threshold of $\widehat{R} \leq 1.05$. As explained in \secref{subsubsec:power_scaling}, only the last sample of the fourth chain of each second step MCMC run is kept for the joint posterior. 

\subsubsection{Results}

We have varied the weighting factor $\beta$ of the two weighted approaches in different increments and will show the most illustrative results here for $\beta \in \{0, 0.1, 0.4, 0.75, 1\}$. We first show posterior predictive plots to build an intuition about the differences between the two approaches, and then report on the evaluation metrics.

\paragraph{Posterior Predictive Plots}

\begin{figure*}[ht]
    \includegraphics[width=\textwidth]{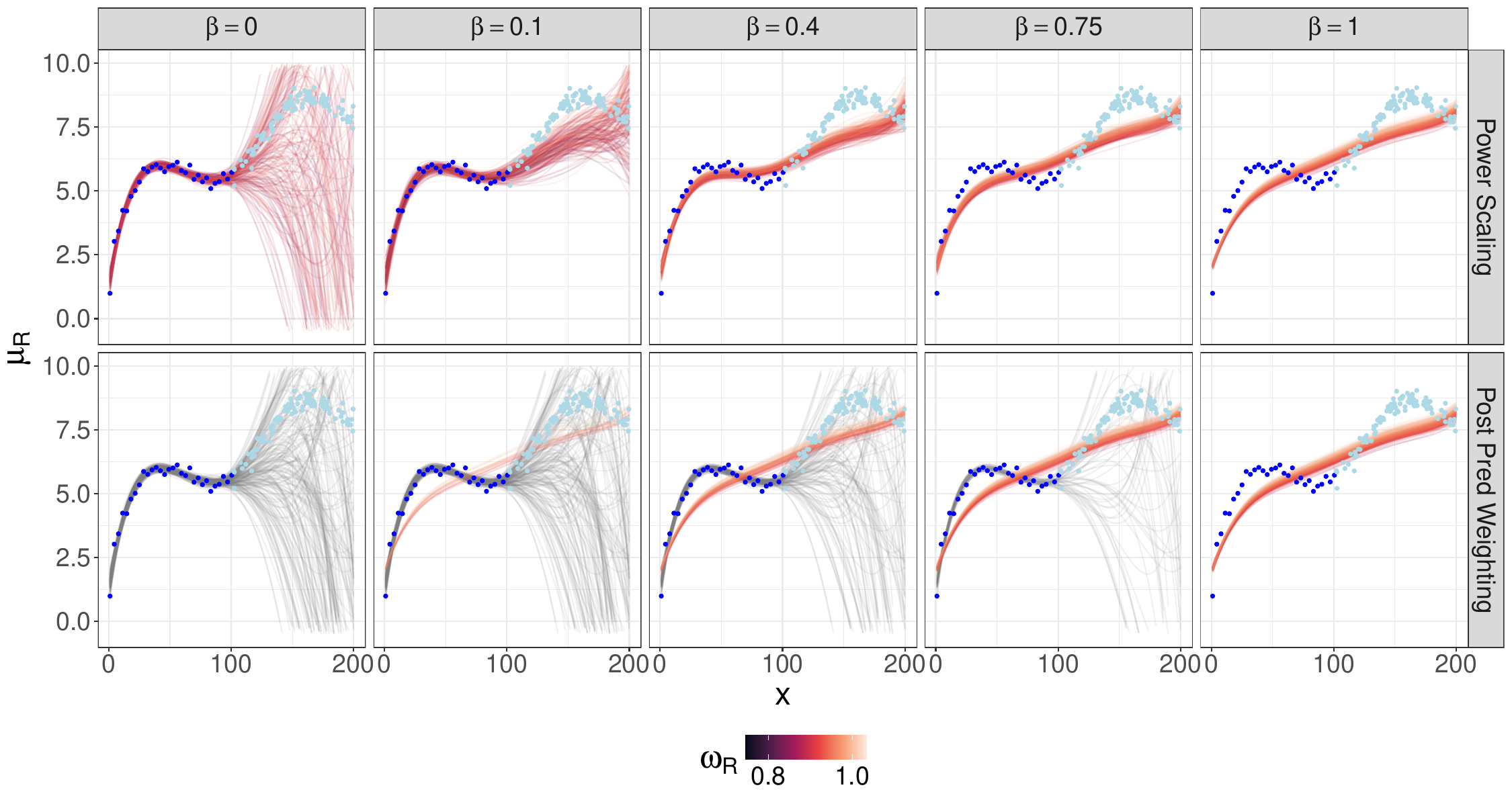}
    \caption{Case study 1: Predictive mean posteriors as obtained from the hybrid surrogates with varied weighting factor $\beta$. Top row: results using the power-scaling approach, bottom row: results from the posterior predictive weighting approach. Each line represents one of 200 posterior samples from the predictive mean posterior distribution, color-coded by the corresponding $\omega_R$ value. The dark and light blue points show train data and out-of-distribution test data, respectively, as also shown in the data setup in \figref{fig:case_study_1_setup}. To enhance clarity, out-of-sample test data are omitted.}\label{fig:case_study_1_posterior_epred}
\end{figure*}

\begin{figure*}[ht]
    \includegraphics[width=\textwidth]{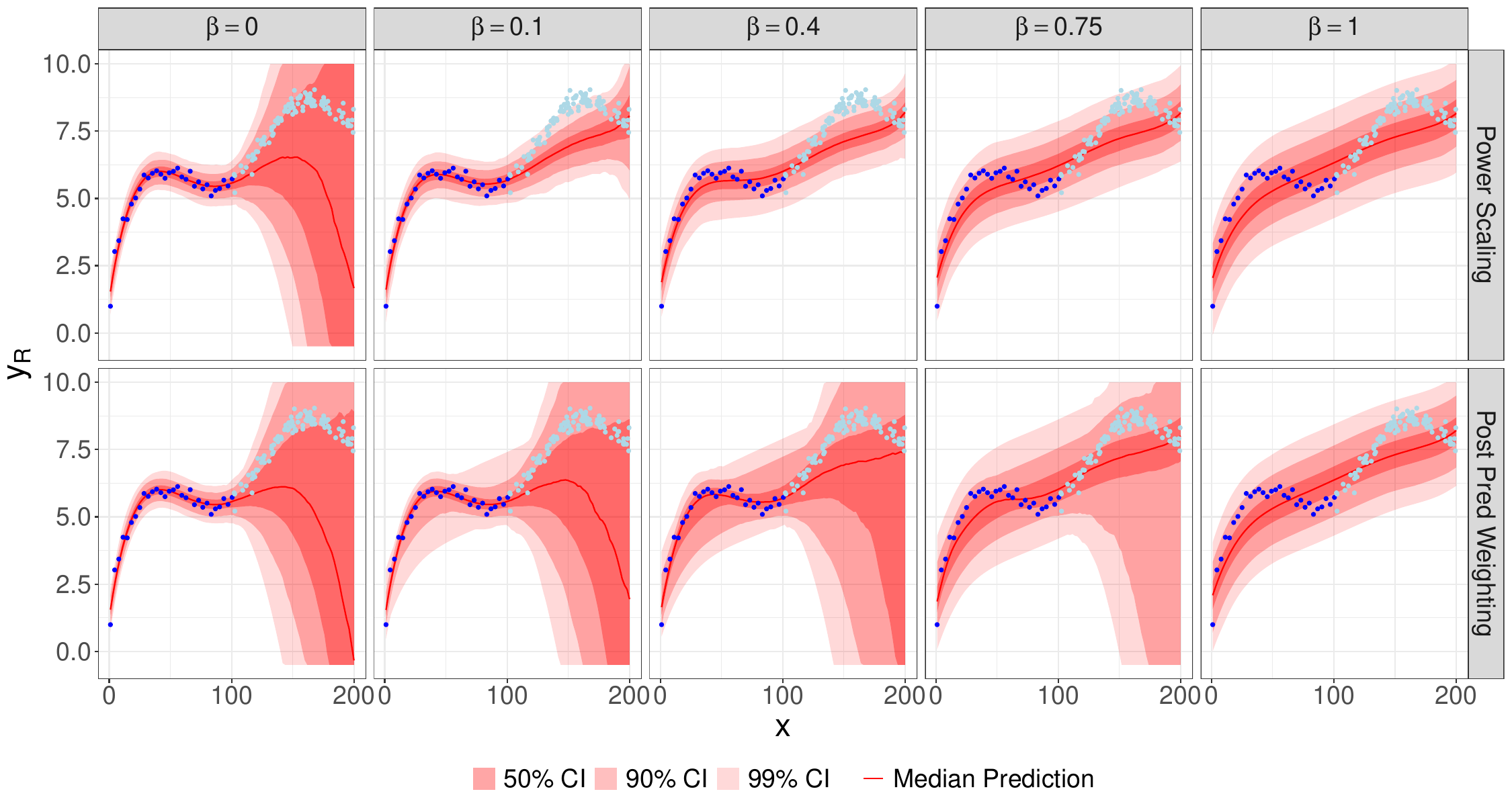}
    \caption{Case study 1: Posterior predictives as obtained from the hybrid surrogates with varied weighting factor $\beta$. Top row: results using the power-scaling approach, bottom row: results from the posterior predictive weighting approach. We display the 50\%, 90\%, 99\% credible intervals along with the median prediction. The dark and light blue points show train data and out-of-distribution test data, respectively, as also shown in the data setup in \figref{fig:case_study_1_setup}.  To enhance clarity, out-of-sample test data are omitted.}\label{fig:case_study_1_posterior_pred}
\end{figure*}

In \figref{fig:case_study_1_posterior_epred}, we show 200 draws of the predictive mean posteriors, which models the underlying mean function. In \figref{fig:case_study_1_posterior_pred}, we display draws from the posterior predictive distribution, which additionally models the noise of the real-world data and the discrepancy induced by the surrogate model in a lumped manner (recall  the definitions of the posterior predictive and the predictive mean posterior distribution in \secref{subsec:evaluation_metrics}). 

For $\beta = 0$, representing the fully data-driven surrogate model, the training data are fitted well but a drastic increase in uncertainty is produced when extrapolating beyond the training range of $x$ with no visible trend. Instead, setting $\beta = 1$ yields the fully simulation-based surrogate model. While this model fits the training data less accurately than the data-driven surrogate, it captures the global trend for out-of-distribution regions better, though it misses the periodic term of the synthetic truth. 

For intermediate $\beta$ values, the power-scaling (PS) surrogate model (top row) tries to match both data sources (with varied relative importance according to $\beta$), such that its predictive posterior seems to ``mediate'' between the simulation-based and data-driven surrogate model. Its uncertainty (or flexibility) increases from $\beta=1$ to $\beta=0$ with reduced impact of the constraints by the simulation data. Yet, it struggles to fit both the OOS and OOD test data. 

In contrast to the PS method, the posterior predictive weighting (PW) method (bottom row of \figref{fig:case_study_1_posterior_epred} and \figref{fig:case_study_1_posterior_pred}), shows a qualitatively different behavior. This method yields a weighted superposition of the simulation-based and data-driven predictive distributions. Note, that the number of gray lines (data-driven surrogate) and red lines (simulation-based surrogate) reflects the proportion to which each posterior predictive distribution has been sampled. The predictive posterior plots for PW over varied $\beta$ enable to visualize the discrepancies that result from using the two different sources of information for surrogate training. These differences can point the modeler to missing or misspecified processes in the simulation model. In our example, the real data would point to a much more uncertain future trend then predicted by the model; this could guide additional analysis into how the simulation model could be improved (asking which processes could exist in the real system that would cause a significant deviation from the trend predicted by the simulation model). On the other hand, the simulation model brings important information about the rising trend that is confirmed by the OOD test data and could be considered when thinking of physics-informed data-driven approaches. 

\begin{figure*}[ht]
    \includegraphics[width=\textwidth]{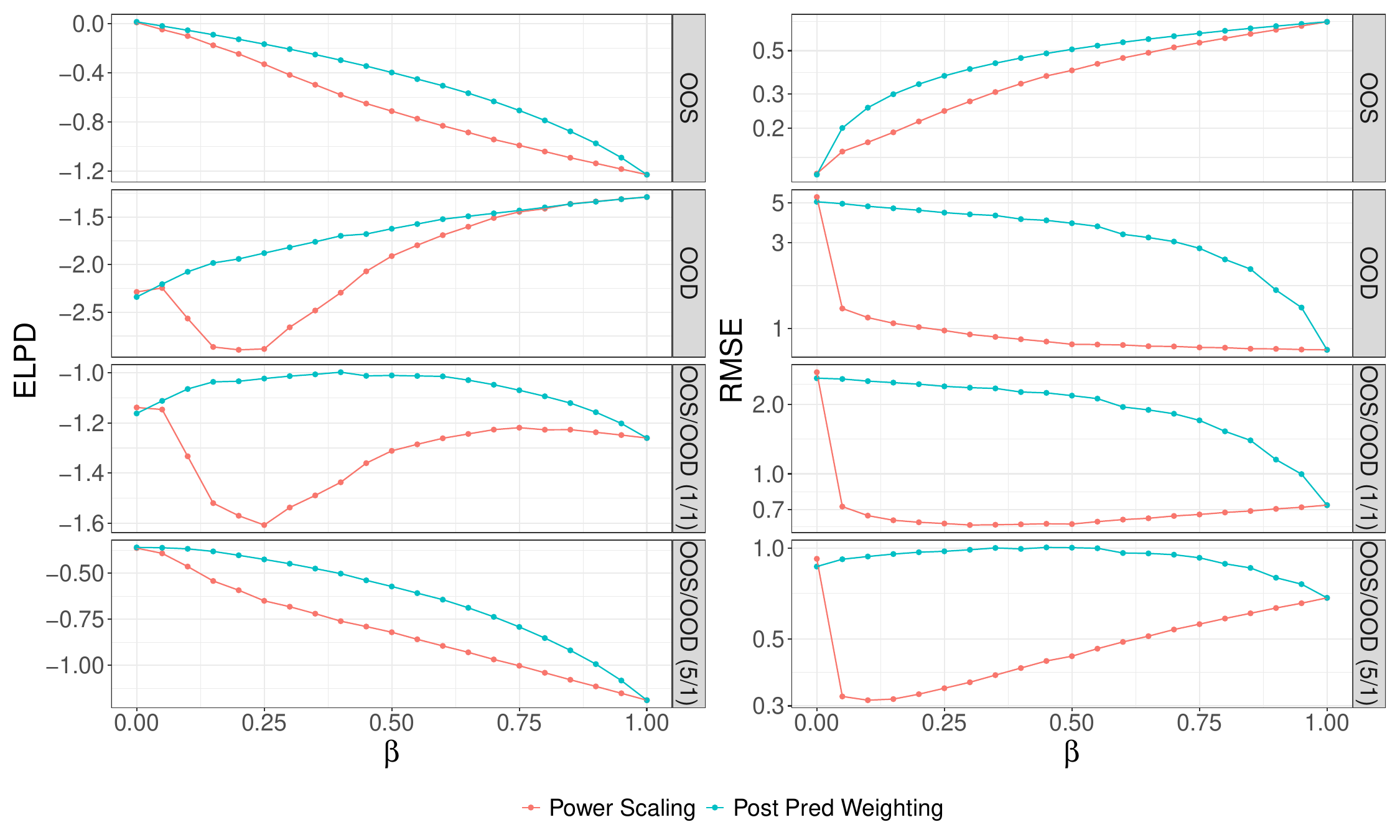}
    \caption{Case study 1: ELPD and RMSE of power-scaling and posterior predictive weighting on test data. The test data splits are out-of-distribution (OOD), out-of-sample in-distribution (OOS), and their combination (OOS/OOD) using 1/1 and 5/1 sample size ratios. For ELPD, higher is better; for RMSE, lower is better.}
    \label{fig:case_study_1_predictive_metrics}
\end{figure*}

\paragraph{Predictive Performance}
To evaluate the predictive capabilities of the two methods quantitatively, we compute the ELPD and the RMSE as explained in \secref{subsec:evaluation_metrics}. Specifically, we evaluate their skill on four different test data scenarios: OOS ($N_R^* = 100$ out-of-sample but in-distribution inputs $x_R \in [1, 100]$), OOD ($N_R^* = 100$ out-of-distribution inputs $x_R \in [100, 200]$), and their combination OOS/OOD ($N_R^*=200$ with equal ratio, 1/1, and $N_R^*=200$ with unequal ratio, 5/1). 

In \figref{fig:case_study_1_predictive_metrics}, we show the ELPD and RMSE metrics for the four scenarios for $\beta \in \{0, 0.05, \dots, 0.95, 1\}$. We observe that for ELPD and RMSE in the OOS setup, optimal performance for both methods is achieved for $\beta = 0$, i.e., with the data-driven surrogate model. This is to be expected because the structure imposed by the simulation model is too rigid to closely mimic the real data. In contrast, the flexible data-driven surrogate can fully exploit the training in this value range and fit the data almost perfectly well. Further, we note that the results for PS and PW do not align perfectly at $\beta = 0$. This discrepancy occurs because in the power-scaling approach $\omega_R$ is sampled from the prior (see \secref{subsubsec:power_scaling}), while the data-driven surrogate model used for posterior predictive weighting only receives the input $x_R$. With finite samples, this leads to minor differences in the results.

In the OOD setup, optimal performance in ELPD and RMSE is achieved with the simulation-based surrogate ($\beta=1$), which is consistent with our previous observation from the posterior prediction plots. This is because the uncertainty of the data-driven surrogate increases drastically for predictions outside its training regime, i.e., it cannot extrapolate with confidence. In contrast, the simulation-based surrogate can use the physical knowledge (in this case the log-linear trend) infused in the training phase through the simulation data. For $0 < \beta < 1$, PW consistently outperforms PS in ELPD, while PS outperforms PW in RMSE. For this particular case study scenario, we can conclude that for OOD data, PW captures the uncertainty better, while the mean prediction produced by PS is closer to the synthetic truth. 

In the third row of \figref{fig:case_study_1_predictive_metrics}, we evaluate the performance on both OOS and OOD test data in a 1/1 ratio. For ELPD, we observe a quadratic curve with an optimum at $\beta \approx 0.4$ for PW, as the weighted superposition of the two predictive distributions leads to an ideal balance between the in-distribution and out-of-distribution performance. Instead for PS, the underestimation of uncertainty in the OOD regime dominates also in the combined OOS/OOD test dataset. While PW outperforms PS for all $\beta$ in ELPD, we observe the opposite situation for RMSE, i.e. PS outperforms PW. Finally, we evaluate the OOS/OOD setup with a 5/1 ratio. Here the quadratic relationship in RMSE is even stronger for PS, with an optimum around $\beta \approx 0.1$. 

Overall, these results suggest that which of the two proposed weighted surrogate training strategies and which relative importance of simulation and real data is optimal depends on the test scenario and the chosen metric. Further,  we discuss in Appendix \secref{appendix:case_study_1} how relatively small changes in the setup can lead to different conclusions. In the considered scenarios, we found that the PS method was better able to recover the underlying synthetic truth, while the PW method was more accurate in modeling the uncertainty. Additionally, visualizing the posterior predictive of PW offers an intuitive tool to understand the potentially competing pieces of information provided by the two data sources.

\subsubsection{Generality of the Proposed Approaches}

To assess the generality of the proposed framework beyond polynomial surrogates, we replicate the setup of case study 1 using a GP as the surrogate model class. Details of implementation as well as results and their discussion are provided in Appendix \ref{appendix:case_study_1_gp_koh}. Overall, the qualitative behavior of the approaches based on GPs mirrors the PCE results. These results confirm that our proposed framework is not restricted to PCE surrogates and, among other possible choices, transfers naturally to GP-based surrogates. 

In addition, we compare the PS and PW approaches against the Kennedy O’Hagan (KOH) calibration framework \citep{kennedyBayesianCalibrationComputer2001}, which represents the canonical GP-based approach for combining simulation and real-world data. The KOH model jointly estimates an emulator GP fitted to the simulation data and an additive discrepancy GP that captures the systematic deviation between the simulator and the real-world data-generating process. Unlike our proposed approaches, KOH does not expose a weighting factor $\beta$ to control the relative influence of the two data sources. Results for case study 1 (Appendix \ref{appendix:case_study_1_gp_koh}) demonstrate that the KOH model produces well-calibrated credible intervals in the OOD region, benefiting from the explicit discrepancy term. Yet, our PS approach outperforms KOH for $\beta \in [0.2, 0.5]$, indicating that an appropriately tuned PS surrogate can recover the underlying mean function more accurately than the KOH discrepancy model in this setting. The comparison with KOH highlights the complementary strengths of the two paradigms: KOH excels at uncertainty quantification through its explicit discrepancy term, while the PS approach can achieve superior mean predictions, without requiring a separate discrepancy model. We emphasize, however, that we do not view our framework as a competitor to KOH. The central contribution of this study is a novel weighting strategy for combining heterogeneous data sources during surrogate training, which operates independently of the chosen surrogate family. The KOH framework, by contrast, focuses on modeling an explicit discrepancy function for GP surrogates.

\subsection{Case Study 2: SIR Model}
\label{subsec:case_study_2}
To demonstrate the applicability to real-world scenarios, we evaluate our weighted surrogate training approach on an epidemiological problem. In the following, we consider the Susceptible-Infected-Recovered (SIR) model as simulation model $\cM(t, \xi, \gamma)$, which is implicitly defined through the following system of ordinary differential equations:
\begin{align}
\begin{split}
    \frac{\mathrm{d}S(t)}{\mathrm{d}t} &= -\xi S(t) \frac{I(t)}{P} \\
    \frac{\mathrm{d}I(t)}{\mathrm{d}t} &= \xi S(t) \frac{I(t)}{P}-\gamma I(t) \\
    \frac{\mathrm{d}R(t)}{\mathrm{d}t} &= \gamma I(t),
\end{split}
\end{align}
where $S(t)$ describes the number of susceptible, $I(t)$ the number of infected, and $R(t)$ the number of recovered individuals at time $t$, and $P$ denotes the constant population. Furthermore, the unknown inputs are $(\xi, \gamma)$, the contact rate and recovery rate \citep{hethcote2000mathinfectios, giordano2020modellingcovid}.

We analyze two different setups: The first is a synthetic setup in which data sources are generated artificially, with an induced misspecification between the real and simulation data. The second setup uses actual measurement data from the early stages of the COVID-19 pandemic.

\subsubsection{Synthetic Setup}
\label{subsubsec:case_study_2_1}

\paragraph{Data Generation}
In the synthetic setup, we generate both simulation and real-world data artificially. The simulation model is based on the SIR model, $\cM(t_S, \xi_S, \gamma_S = 0.55)$, where the recovery rate is fixed at $\gamma_S = 0.55$. The considered output is the number of infected $I_S$, which is typically observed as measurement data. For the real data-generating process $\cG$, we introduce a misspecification between the simulator and the ``real-world'' data through different recovery rates. Specifically, we use the SIR model with contact rate $\xi_R = 1.6$ and recovery rate $\gamma_R = 0.7$. Again, we consider as output the number of infected $I_R$. To generate over-dispersed real-world count data, we sample from a negative binomial distribution with mean and precision parametrization (see Appendix \secref{appendix:negbinom_lognormal}): 
\begin{align}
    y_R \sim \text{NegBinomial}(I_R, \sigma_\epsilon = 5).
\end{align}

To generate simulation training data, we consider the first $N_S = 100$ points of the 2-dimensional Sobol sequence \citep{sobol1967distributionpointscube}, scaled to the ranges $[1, 14]$ and $[1, 3]$ for the inputs $t_S$ and $\xi_S$, respectively. These inputs are fed into the simulation model $\cM$ to generate the outputs, the number of infected $I_S$. For real-world training data, we use $N_R = 30$ evenly spaced input points in the first half of the simulation input space, i.e. $t_R \in [1, 7.5]$. The real-world number of infected $y_R$ are generated using $\cG$ at the real-world inputs and $\sigma_\epsilon = 5$. 

\paragraph{Hybrid Surrogate Modeling}
As surrogate we consider again a PCE, as described in case study 1 (see \secref{subsec:case_study_1}). The surrogate takes the two dimensional input $(t, \xi)$, and models as output the number of infected $I(t)$:
\begin{align}
    \widetilde{\cM}(t, \xi, c) &= \sum_{i=0}^{d-1} c_i \psi_i(t, \xi).
\end{align}
We use two-dimensional Legendre polynomials up to a degree of 5 and follow the standard truncation scheme, leading to d = 21 polynomials in total. 
When fitting the models, the inputs $(t, \xi)$ are linearly scaled to the standard scaling of the Legendre polynomials, i.e. $[-1, 1]$. 
We specify the following probabilistic model for the power-scaling approach:
\begin{align}
    I_S &\sim \text{Log-Normal}(\widetilde{\cM}(t_S, \xi_S, c), \sigma)^{\alpha_S}\\
    y_R &\sim \text{Log-Normal}(\widetilde{\cM}(t_R, \xi_R, c), \sigma)^{\alpha_R}\\
    \widetilde{\cM}(t, \xi, c) &= \sum_{i=0}^{d-1} c_i \psi_i(t, \xi) \\
    \xi_R &\sim \text{Normal}_{[1, 3]}(2, 0.5)\\
    \sigma &\sim \text{Half-Normal}(0.5)\\
    c_i &\sim \text{Normal}(0, 5).
\end{align}

To avoid exact zero responses, which would lead to negative infinity when evaluating the log-normal distribution, we added $1$ to the outputs $I_S$ and $y_R$.
Although the real data was generated using a negative binomial distribution, we used log-normal surrogate likelihoods. This choice aligns better with the simulation data $I_S$, which is continuous and zero-bounded. While real-world data are discrete, the log-normal likelihood ensures maximal utilization of simulation information. Using a negative binomial likelihood would require noisy simulation data, i.e. $y_S$ sampled from a negative binomial. This would significantly increase the uncertainty in the simulation data which would reduce the efficiency of the surrogate training process. Since the error parameter $\sigma$ is shared across simulation and real-world data, the same likelihood family must be used for both sources. 

For the posterior predictive weighting method, we fit a simulation-based and data-driven model using the respective likelihoods and priors specified above. We sample from the respective implied posteriors using HMC via Stan with the same specifics as described in \secref{subsec:case_study_1}.

\begin{figure*}[ht]
    \includegraphics[width=\textwidth]{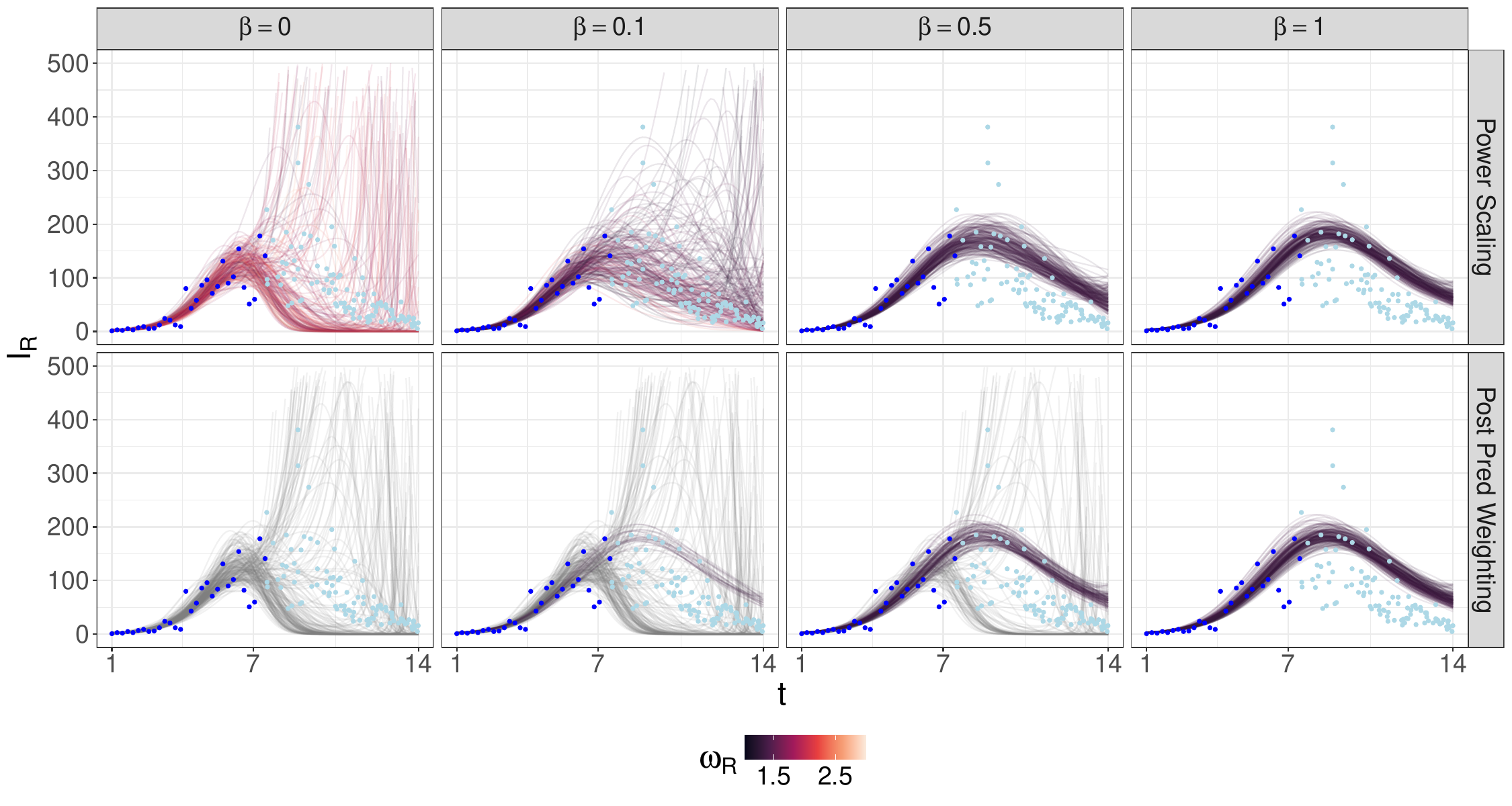}
    \caption{Case study 2.1: Predictive mean posteriors as obtained from the hybrid surrogates with varied weighting factor $\beta$. Top row: results using the power-scaling approach, bottom row: results from the posterior predictive weighting approach. Each line represents one of 200 posterior samples from the predictive mean posterior distribution, color-coded by the corresponding $\omega_R$ value. The dark and light blue points show train data and out-of-distribution test data, respectively}\label{fig:case_study_2_1_posterior_epred}
\end{figure*}

\begin{figure*}[ht]
    \includegraphics[width=\textwidth]{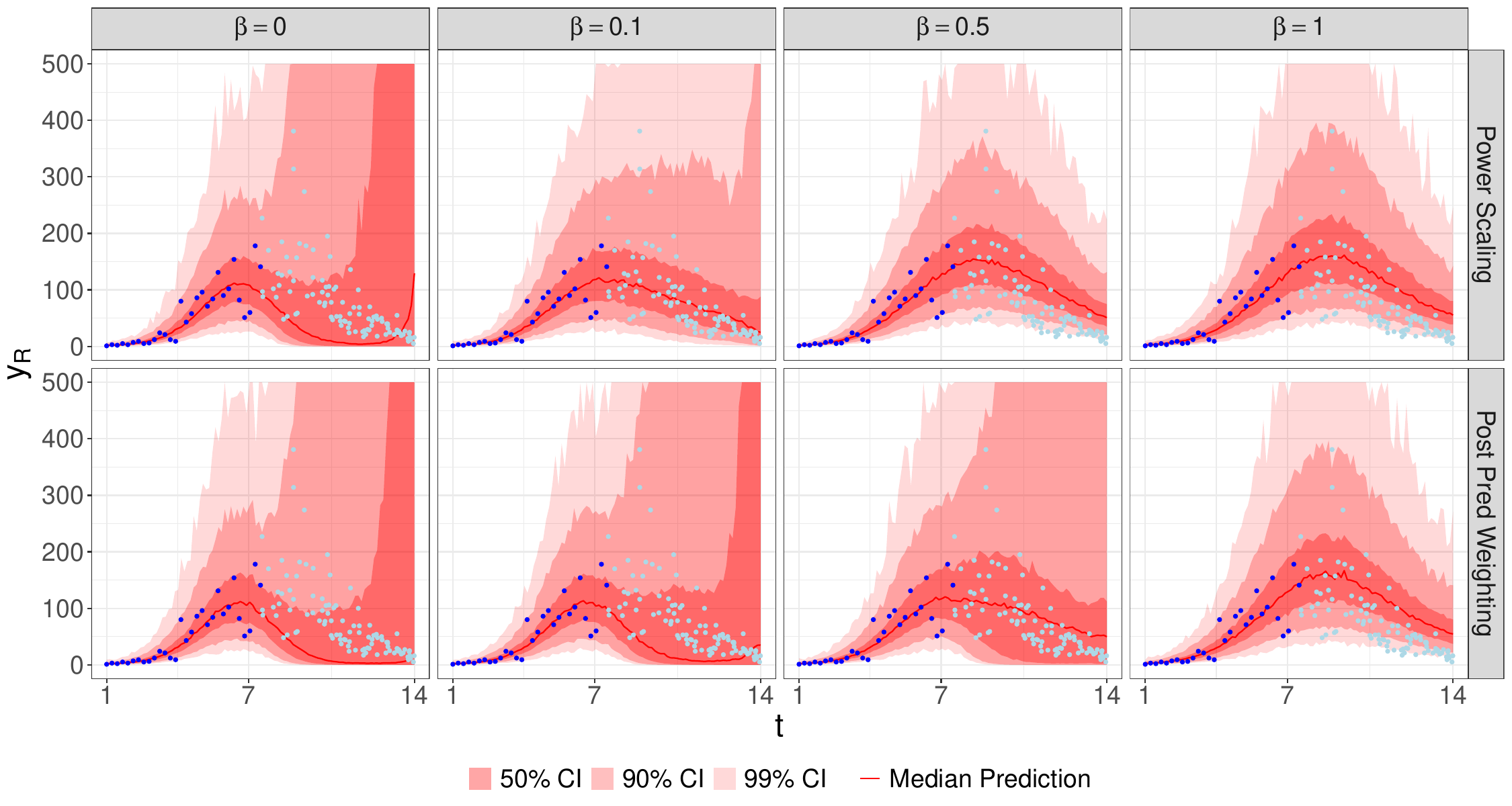}
    \caption{Case study 2.1: Posterior predictives as obtained from the hybrid surrogates with varied weighting factor $\beta$. Top row: results using the power-scaling approach, bottom row: results from the posterior predictive weighting approach. We display the 50\%, 90\%, 99\% credible intervals along with the median prediction. The dark and light blue points show train data and out-of-distribution test data, respectively.}\label{fig:case_study_2_1_posterior_pred}
\end{figure*}

\paragraph{Posterior Predictive Plots}
\figref{fig:case_study_2_1_posterior_epred} illustrates the predictive mean posterior and \figref{fig:case_study_2_1_posterior_pred} the posterior predictions for the weighting factors $\beta \in \{0, 0.1, 0.5, 1\}$.

The predictive mean posteriors of $I_R$ in \figref{fig:case_study_2_1_posterior_epred} represent the surrogate’s approximation to the underlying synthetic truth (noise-free real-world data-generating process). For $\beta = 0$, corresponding to the fully data-driven surrogate, the training regime is accurately modeled. However, the mean predictions for the out-of-distribution (OOD) region are highly uncertain, with some samples near zero and other samples yielding implausibly large predictions. For $\beta = 1$, representing the fully simulation-based surrogate, the mean predictions align with typical infection curves constrained by the simulation. However, in the OOD region, the decline in infections is slower than observed in real data due to the simulation misspecification. For intermediate values of $\beta$ using power-scaling (PS), the mean predictions at $\beta = 0.1$ align well with observed data but remain uncertain, while at $\beta = 0.5$, the mean predictions exceed the real data in the OOD region. Using posterior predictive weighting (PW), intermediate values of $\beta$ show a weighted superposition of data-driven and simulation-based predictions. This is because, to calculate a weighted average of sample-based posterior predictive distributions, the ratio of samples chosen from the respective predictive distributions is set according to the weighting factor $\beta$. The superposition nicely visualizes the competing information drawn from the two data sources: while the real-world data favor an earlier and steeper decline, the simulation data favor a slower decline and a stabilization at a higher level of infected individuals. These insights lead to further diagnostic questions, like: what could cause the different ``messages'' of data and simulation? How confident are we about the two sources of information? Are there any additional data or expert knowledge that could help decipher which projected trend is more realistic?

The posterior predictions of $y_R$ in \figref{fig:case_study_2_1_posterior_pred} capture the actual noisy real-world data. At $\beta = 0$, the fully data-driven surrogate models training data and its uncertainty well. However, in the test regime, predictions show rapidly increasing uncertainty, with median values trending toward zero. At $\beta = 1$, the simulation-based surrogate captures uncertainty well in both training and test regimes. Yet, similar to the mean predictions that are far from the training range, the overall prediction is too high. For intermediate $\beta$ values using power-scaling, predictions at $\beta = 0.1$ effectively balance real and simulation data influences. However, at $\beta = 0.5$, the increasing simulation influence results in slightly narrower but more biased predictive intervals. For posterior predictive weighting, predictions reflect a strong influence from the data-driven surrogate, as can be seen by the overly broad credible intervals for both values of $\beta$, namely 0.1 and 0.5.

\paragraph{Predictive Performance}

\begin{figure*}[ht]
    \includegraphics[width=\textwidth]{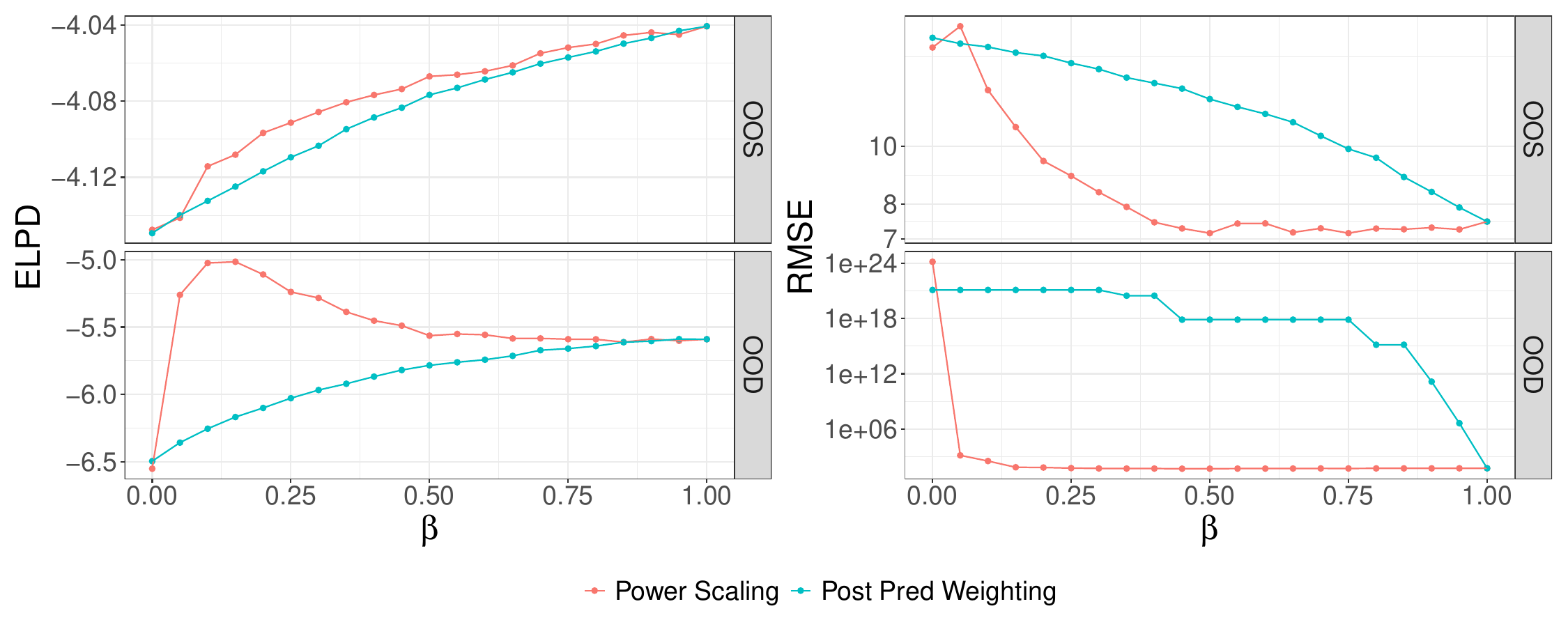}
    \caption{Case study 2.1: ELPD and RMSE of power-scaling and posterior predictive weighting on test data. The test data splits are out-of-sample in-distribution (OOS) and out-of-distribution (OOD). For ELPD, higher is better; for RMSE, lower is better.}
    \label{fig:case_study_2_1_predictive_metrics}
\end{figure*}

To evaluate the predictive capabilities of the approaches quantitatively, we calculate the ELPD and RMSE (for details, see \secref{subsec:evaluation_metrics}). Specifically, we evaluate two test scenarios, differing in whether they are in training distribution or outside of it: OOS ($N_R^* = 100$ out-of-sample but in-distribution inputs $t_R \in [1, 7.5]$) and OOD ($N_R^* = 100$ out-of-distribution inputs $t_R \in [7.5, 14]$). In \figref{fig:case_study_2_1_predictive_metrics}, we present the ELPD and RMSE metrics for $\beta \in \{0, 0.05, \dots, 0.95, 1\}$ across both scenarios. Most of the variation between the two approaches and $\beta$ happens in the OOD scenario. Therefore, we omit for this case study the combination of OOS and OOD since the results are very similar to the OOD scenario.

For OOS, ELPD results show that the simulation-based surrogate ($\beta = 1$) achieves the best performance, while both PW and PS show linear improvements with increasing $\beta$. For RMSE, the simulation-based surrogate also performs best, with PW showing steady improvements and PS achieving faster improvements up to $\beta = 0.5$ before plateauing.

For OOD, ELPD results show that PS achieves maximum performance at $\beta \approx 0.1$, outperforming both the data-driven surrogate ($\beta = 0$) and the simulation-based surrogate ($\beta = 1$). In contrast, PW shows a monotonic improvement in ELPD as $\beta$ increases but does not reach the peak performance of PS. RMSE values for $\beta = 0$ are extremely high due to occasional large predictions from the data-driven surrogate. PW improves slowly, while PS quickly stabilizes from around $\beta \approx 0.25$.

For this scenario, OOS evaluations reveal similar performance trends across both approaches (PW and PS). However, for OOD, PS outperforms PW by leveraging earlier regularization from the simulation data, effectively avoiding extreme predictions.

\subsubsection{Covid19 Real Data Setup}
In the second setup, we use actual real-world measurement data from the early stages of the COVID-19 pandemic. Specifically, we utilize publicly available COVID-19 data for Italy from \cite{guidotti2020, guidotti2022}, focusing on the first wave (February 24 to May 7, 2020). The objective is to predict future infection numbers by applying our weighted surrogate approaches, leveraging both simulation and real-world data for improved predictive accuracy.

\paragraph{Data}
As simulation model, we again use the SIR model but this time with one known input and two unknown free input parameters: $\cM(t_S, \xi_S, \gamma_S)$. To generate simulation training data, we consider the first $N_S = 1,000$ points of the 3-dimensional Sobol sequence \citep{sobol1967distributionpointscube} scaled to the ranges $[0.1, 1.3]$, $[1, 3]$, and $[0.1, 1.0]$ for the inputs $t_S$, $\xi_S$, and $\gamma_S$, respectively. These inputs are then passed through the simulation model $\cM$ to generate the outputs, the number of infected $I_S$. 

For real-world training data, we use $N_R = 60$ daily reported data from February 24 to April 23, 2020. We calculate the number of infected $y_R$ by subtracting the number of recovered from the total number of confirmed cases. By fixing the population to $P = 60,421,760$ (based on data from \cite{guidotti2022}), the data can be rescaled to cases per $10^5$ people.

\paragraph{Hybrid Surrogate Modeling}
In this setup, the input to the surrogate is three-dimensional $(t, \xi, \gamma)$. We use a PCE surrogate with a maximum degree of polynomials of 5, resulting in $d = 56$ polynomials. The probabilistic model is defined as follows:
\begin{align}
    I_S &\sim \text{Log-Normal}(\widetilde{\cM}(t_S, \xi_S, \gamma_S, c), \sigma)^{\alpha_S}\\
    y_R &\sim \text{Log-Normal}(\widetilde{\cM}(t_R, \xi_R, \gamma_R, c), \sigma)^{\alpha_R}\\
    \widetilde{\cM}(t, \xi, \gamma, c) &= \sum_{i=0}^{d-1} c_i \psi_i(t, \xi, \gamma) \\
    \xi_R &\sim \text{Normal}_{[0.1, 1.3]}(0.3, 0.1)\\
    \gamma_R &\sim \text{Normal}_{[0.1, 1.0]}(0.2, 0.1)\\
    \sigma &\sim \text{Half-Normal}(0.5)\\
    c_i &\sim \text{Normal}(0, 5).
\end{align}
Compared to the first setup, we added a prior on $\gamma$ and changed the prior of $\xi$ to address the new setup requirements and avoid identification issues. 
HMC via Stan is again used for posterior sampling, with specifications matching those used in the previous setup.

\begin{figure*}[ht]
    \includegraphics[width=\textwidth]{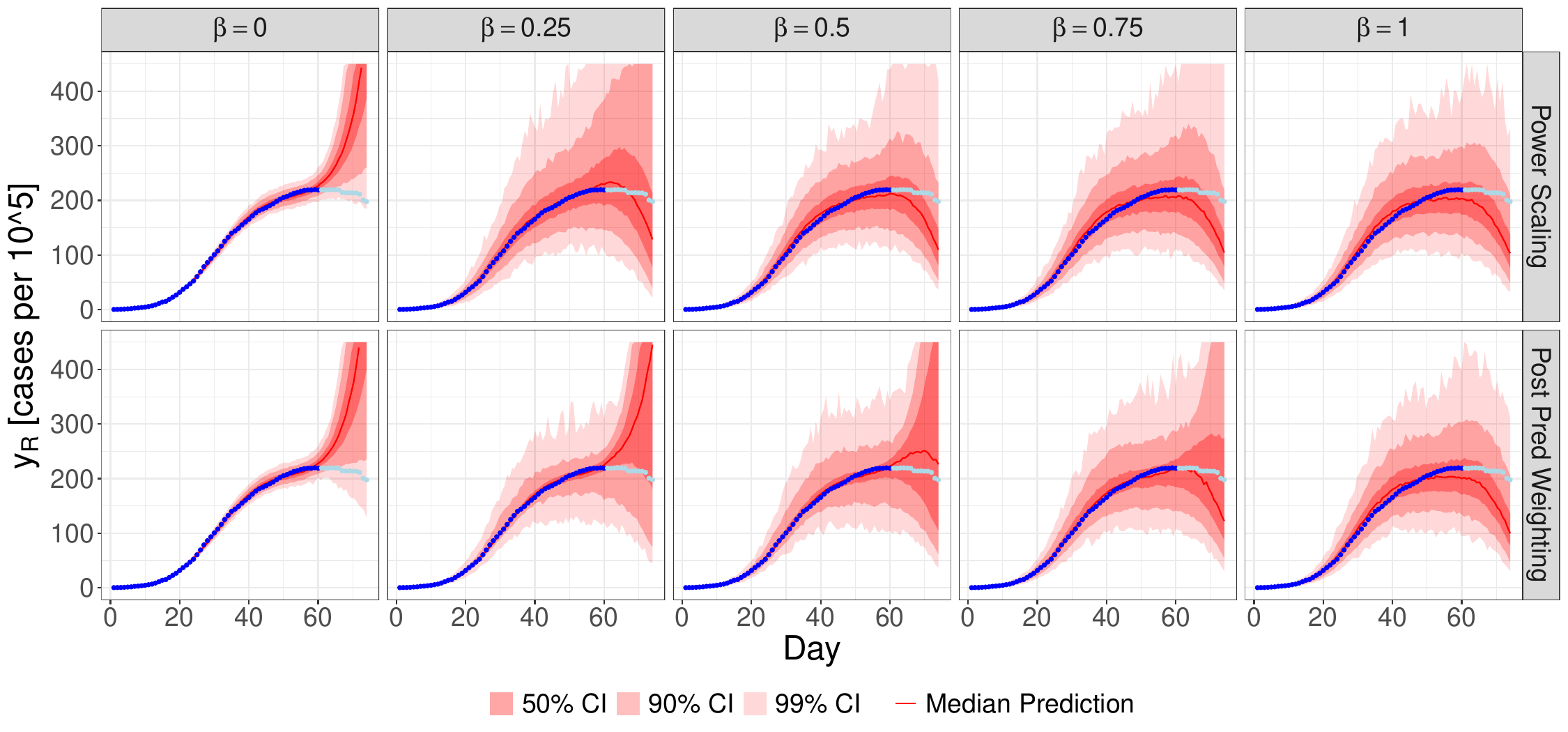}
    \caption{Case study 2.2: Posterior predictives as obtained from the hybrid surrogates with varied weighting factor $\beta$. Top row: results using the power-scaling approach, bottom row: results from the posterior predictive weighting approach. We display the 50\%, 90\%, 99\% credible intervals along with the median prediction. The dark and light blue points show train data and out-of-distribution test data, respectively.}\label{fig:case_study_2_2_posterior_pred}
\end{figure*}

\paragraph{Posterior Predictive Plots}
\figref{fig:case_study_2_2_posterior_pred} illustrates the posterior predictions for the weighting factors $\beta \in \{0, 0.25, 0.5, 0.75 ,1\}$. In contrast to the synthetic scenarios, we cannot compare mean predictions to a synthetic truth since it is not available for actual measurement data. Consequently, we focus on the posterior predictive plots. The data-driven surrogate model, as in previous case studies, models the training data well. However, it fails extrapolating, predicting an increase in infections despite a decline in the actual number of infections. Instead, the simulation-based surrogate has a rather high uncertainty on the training data regime. This is due to the simplifying assumptions in the SIR model (e.g., a constant contact rate $\xi$ over time \citep{schmidt2021probabilisticstatespace}), which results in a higher approximation error and consequently, a high uncertainty in the posterior predictive. For OOD data, the simulation-based surrogate predicts the other extreme, a too strong decrease in infections. For power-scaling, intermediate $\beta$ values lead to wider uncertainty bands, balancing the two data sources. For PW, we observe that for $\beta = 0.5$ the 50\% credible intervals entail both the simulation-based and data-driven surrogate prediction. However, there is low actual posterior probability mass at the test data points, since both predictions on their own lie in very different value ranges.

\paragraph{Predictive Performance}

\begin{figure*}[ht]
    \includegraphics[width=0.5\textwidth]{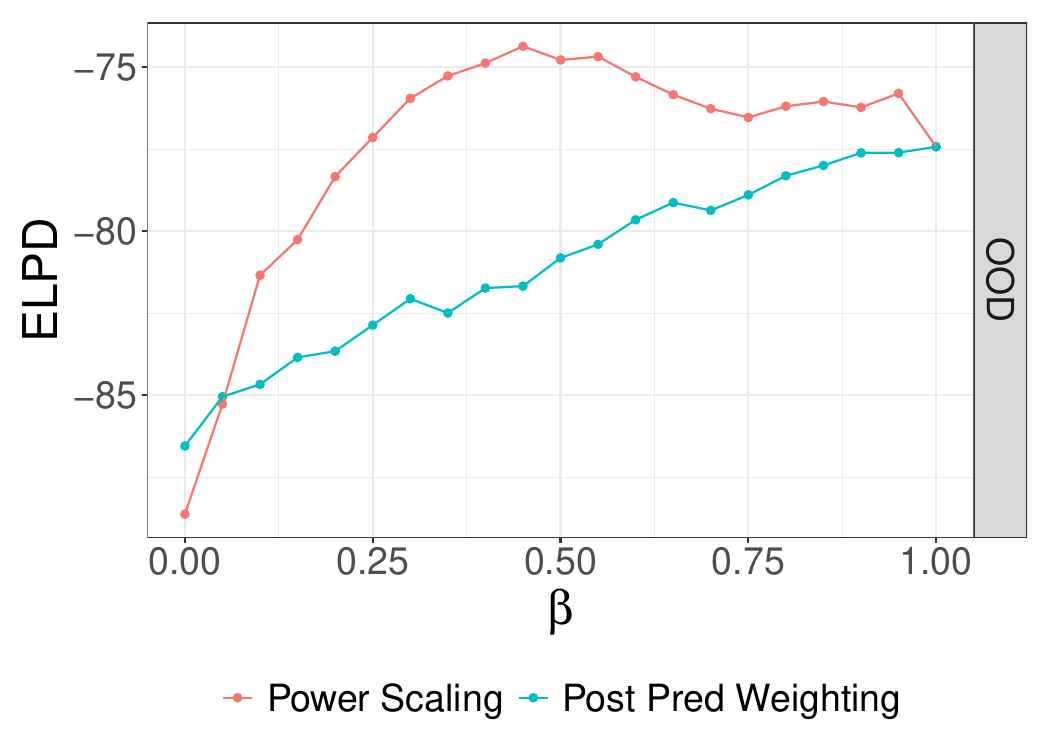}
    \centering
    \caption{Case study 2.2: ELPD and RMSE of power-scaling and posterior predictive weighting on test data. The test data consists of 14 unseen future days.}
    \label{fig:case_study_2_2_predictive_metrics}
\end{figure*}

To evaluate the predictive capabilities of the approaches quantitatively, we calculate the ELPD for the future test data, which consists of 14 days of infection counts (from April 24 to May 7, 2020). However, since no synthetic truth is available, we cannot compute the RMSE from the ground truth. In \figref{fig:case_study_2_2_predictive_metrics}, we present the ELPD metric for $\beta \in \{0, 0.05, \dots, 0.95, 1\}$. We observe that the data-driven model performs worse than the simulation-based model. For the PW method, performance improves monotonically with increasing $\beta$. Power-scaling achieves an overall optimum at $\beta \approx 0.45$. 

\section{Summary, Conclusions \& Outlook}
\label{sec:conclusion}

To address the gap between simulation and reality in surrogate models, we proposed a general-purpose framework for training Bayesian surrogate models in the spirit of hybrid modeling, that combines multiple data sources (simulation and real-world data). Within this framework, we introduced two fundamentally different approaches in order to combine multiple data sources in the training step of a surrogate. Additionally, it allows for the user to assign a weight to each data source, and thereby control their relative influence.

We demonstrated the utility of our framework by applying it to both synthetic and real-world scenarios. Concretely, we showed that the weighted multi-source approach can improve predictive performance compared to standard single-source surrogate training approaches. Additionally, the approaches provide insights about system behavior or potential model deficits.

Our case studies include a synthetic example and real-world examples from epidemiology. We focused on scenarios where the simulator is misspecified relative to real-world processes. In these contexts, our weighted approaches have the potential to improve surrogate predictive performance, particularly in regions outside the training data. However, the relative performance of the "pure" (simulation-based or data-driven) surrogates vs. the hybrid surrogates depends on the evaluation metric and the specific case study context.

Since our framework allows to seamlessly scan this ``surrogate space'' by varying the weighting factor from 0 (data-driven) to 1 (simulation-based), we recommend to perform this analysis on the case study at hand, to compare performances and gain valuable insights into the information content of the different sources of information for surrogate training. Our results indicate that the relative accuracy of the weighted approaches in modeling uncertainty varies depending on the scenario. For our synthetic scenarios, we additionally measured how well the methods can recover the underlying synthetic truth. Here we found that the power-scaling method consistently outperforms the posterior predictive weighting method. This is because in the latter, poorly constrained data-driven surrogates can produce unreliable predictions, which persist in the final weighted output. In contrast, the power-scaling approach integrates data sources during training, leveraging simulation data as a regularizer, thereby avoiding this issue. Yet, visualizing the posterior predictive distributions as by posterior predictive weighting offers an intuitive way to interpret and communicate the potentially competing contributions of the two data sources. These visualizations can also help identify missing or misspecified processes in the simulation model, providing valuable insights for model refinement. Overall, these results highlight the strengths and trade-offs of the two proposed methods for weighted hybrid surrogate training. We therefore recommend to pick each method according to the analysis goal (prediction skill vs. diagnostic insight), at best combining the strengths of both approaches if computationally feasible.

Regarding computational and implementation effort, the two methods differ. The implementation of predictive weighting is relatively straightforward, involving the independent training of two surrogates on different data sources, followed by combining their predictions. However, for a given weighting factor, it does not yield a single joint model that can perform inference on unknown parameters. In contrast, power-scaling trains a single surrogate model, which integrates both data sources and produces distinct posteriors for unknown parameters for each weighting factor. While this offers a unified modeling framework, it is more complex to implement and more prone to identifiability issues.

One partially open challenge concerns finding the optimal weighting factor. In this paper, we performed a simple line search over the unit interval. This also helped illustrating the general implications of the weighting factor.  
However, for the power-scaling approach, this may quickly become computationally infeasible, since a new surrogate model has to be trained from scratch for each considered weight. As a potential remedy, one could use importance sampling \citep{vehtariSurveyBayesianPredictive2012} or moment matching \citep{paananen2021implicitlyadaptiveimportance} to reduce the number of surrogate refits; methods that have been applied successfully in other contexts involving power-scaled models \citep{kallioinenDetectingDiagnosingPrior2024}.
For the posterior predictive weighting approach, finding the optimal weights can be performed after model training using standard model averaging methods, for example, stacking of posterior predictive distributions \citep{yaoUsingStackingAverage2018}. As such, for this approach, the computational costs of finding the optimal weights are usually low compared to the surrogate training.

While we mainly used Polynomial Chaos Expansions as surrogates, the proposed framework is not relying on specific surrogate models (and we have demonstrated its applicability to Gaussian Processes), provided they support sampling from posterior (predictive) distributions. However, for computationally intensive surrogates, such as neural networks with millions of parameters, adjustments would be necessary. For instance, approximate methods for posterior estimation, such as last-layer variational inference \cite{Kristiadi2020BeingBayesian, fiedler2023improveduncertaintyquantification, harrison2023variationalbayesianlast} could be used instead of computationally expensive MCMC algorithms.

Although this study focused on two data sources (one simulation- and one real-world-based source), the framework can be extended to integrate more than two data sources, each with its own weight. This flexibility is valuable in scenarios where multiple simulation models, differing in complexity and plausibility, must be balanced against real-world measurements, and/or in scenarios with multiple real-world data sets of various fidelity. Assigning relative weights could enable more nuanced integration of diverse information.

Another promising future direction is to extend the framework toward input-dependent weighting or error modeling functions. Allowing the relative influence of simulation and real-world data to vary across the input space could further improve predictive performance, especially in heterogeneous or partially observed regimes.

In summary, we presented a flexible framework for weighting multiple data sources in Bayesian surrogate model training. Our framework enables the creation of uncertainty-aware surrogate models with improved predictive capabilities for real-world phenomena. The insights gained from our approach can then be utilized for future improvements in simulation or hybrid model development.

%% file: content/acknowledgments.tex
Partially funded by Deutsche Forschungsgemeinschaft (DFG, German Research Foundation) under Germany’s Excellence Strategy EXC 2075 -- 390740016 and Collaborative Research Center 391 (Spatio-Temporal Statistics for the Transition of Energy and Transport) -- 520388526. We acknowledge the support by the Stuttgart Center for Simulation Science (SimTech).

%% file: content/appendix.tex
\section{Notation}

\begin{table}[ht]
  \caption{Notation used in the paper.}
  \centering
  \renewcommand{\arraystretch}{1}
  \begin{tabular}{cl}
    \toprule
    \textbf{Symbol}            & \textbf{Description} \\
    \midrule
    $\cM$                      & simulation model \\
    $\cD_S$                    & simulation training data \\
    $x_S$                      & (known) simulation input variables \\
    $\omega_S$                 & (unknown) simulation input parameters \\
    $y_S$                      & simulation output \\ 
    $N_S$                      & number simulation training data points \\ 
    \hdashline \\[-1.9ex] 
    $\cG$                      & real-world data-generating process \\
    $\cD_R$                    & real-world data \\
    $x_R$                      & known real-world measurement \\
                               & input variables \\
    $\omega_R$                 & unknown real-world measurement \\
                               & input parameters \\
    $y_R$                      & real-world measurement output \\
    $N_R$                      & number real-world data points \\
    $\sigma_\epsilon$          & measurement noise parameter \\ 
    \hdashline \\[-1.9ex] 
    $\widetilde{\cM}$          & surrogate model \\
    $c$                        & surrogate parameters \\
    $\beta$                    & weighting factor \\
    $\alpha_S$                 & scaling factor (simulation data) \\
    $\alpha_R$                 & scaling factor (real-world data) \\
    $\sigma_S$                 & simulation approximation error parameter \\ 
    $\sigma_R$                 & real-world approximation error parameter \\ 
    $\sigma$                   & (lumped) approximation error parameter \\ 
    \bottomrule
  \end{tabular}
  \label{table:variable_glossary}
\end{table}

\section{Negative binomial and log-normal distributions}
\label{appendix:negbinom_lognormal}
Below, we show the negative binomial and log-normal distributions. 

The probability mass function of the negative binomial distribution in the mean-precision parametrization for scalar count $n \in \nN$ with the two positive parameters $\mu \in \nR^+$ and $\phi \in \nR^+$ is given by:
\begin{align}
\begin{split}
    &p_{\text{NegBinomial}}(n \mid \mu, \phi) \\
    &=\binom{n + \phi - 1}{n} \left( \frac{\mu}{\mu + \phi} \right)^n \left( \frac{\phi}{\mu + \phi} \right)^\phi.
\end{split}
\end{align}
In this parametrization the mean and variance are given by
\begin{align}
    \nE[n] = \mu \quad \text{and} \quad \text{Var}[n] = \mu + \frac{\mu^2}{\phi}.
\end{align}
The probability density function of the log-normal distribution for a positive scalar $y \in \nR^+$ with the parameters $\mu \in \nR$ and $\sigma \in \nR^+$ is given by:
\begin{align}
\begin{split}
    &p_{\text{Log-Normal}}(y \mid \mu, \sigma) \\
    &=\frac{1}{\sqrt{2 \pi} \sigma} \frac{1}{y} \exp \left( - \frac{1}{2} \left( \frac{\log y - \mu }{\sigma} \right)^2 \right).
\end{split}
\end{align}
In this parameterization the mean and variance are given by
\begin{align}
    &\nE[y] = \exp \left(\mu + \frac{\sigma^2}{2} \right) \quad \text{and} \\
    &\text{Var}[y] = [\exp(\sigma^2) - 1] \exp(2 \mu + \sigma^2).
\end{align}

\section{Case Study 1}
\label{appendix:case_study_1}

\subsection{Real-world parameter posterior distributions}
\begin{figure*}[ht]
    \includegraphics[width=\textwidth]{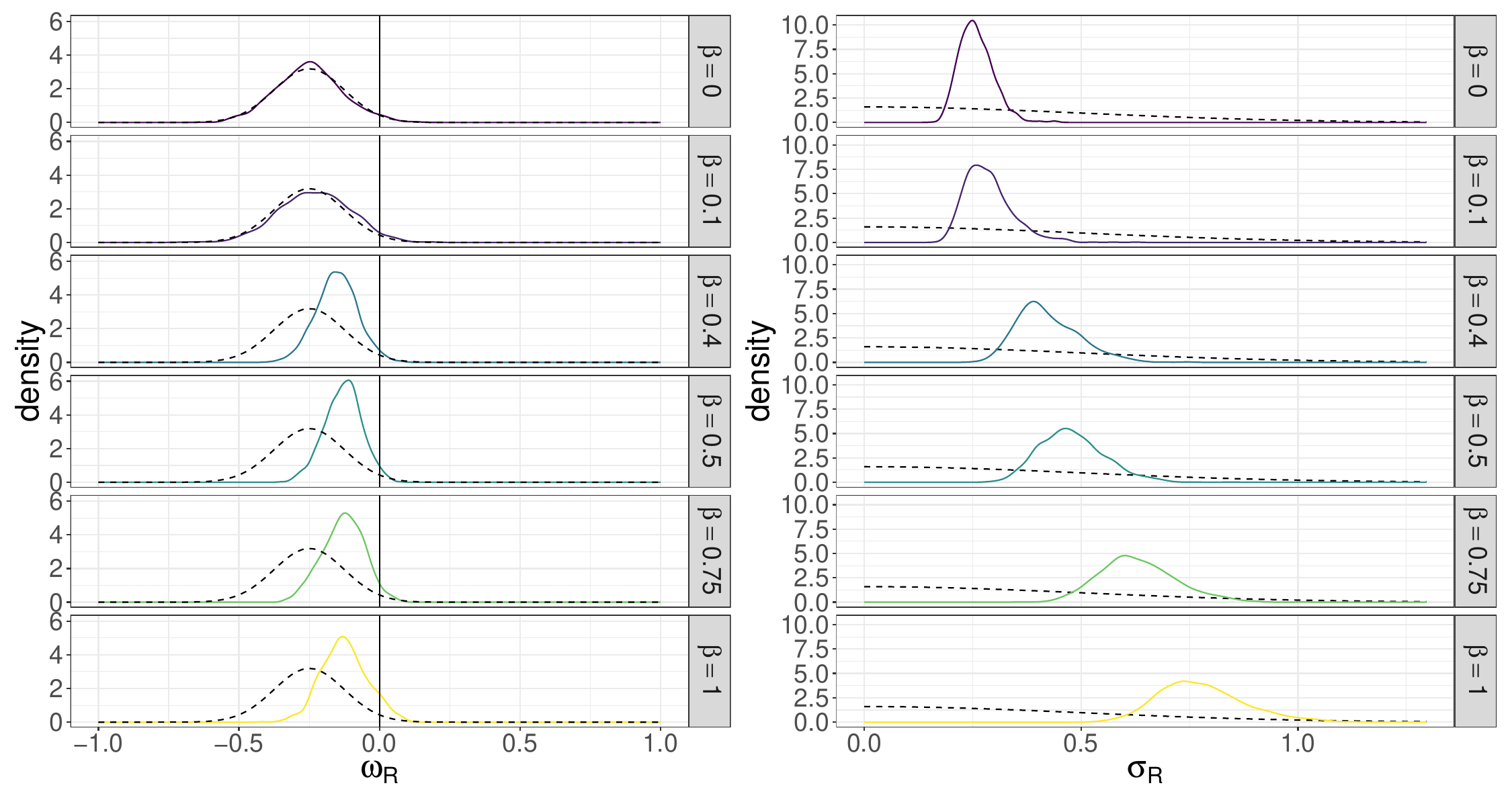}
    \caption{Case study 1: Posterior densities of $\omega_R$ and $\sigma_R$ using the power-scaling method for $\beta = \{ 0, 0.1, 0.4, 0.5, 0.75, 1\}$. Left column: posterior $p^\beta_{\rm ps}(\omega_R \mid \cD_R)$, right column: posterior $p^\beta_{\rm ps}(\sigma_R \mid \cD_R)$. Each posterior density is colored by its $\beta$ value. The dashed lines depict the prior density. The vertical line on the left column depicts the true value $\omega_R^*$.}
    \label{fig:case_study_1_omega_sigma_posterior}
\end{figure*}

In the following, we show additional plots for the setup described in case study 1 (see \ref{subsec:case_study_1}). In \figref{fig:case_study_1_omega_sigma_posterior} we depict the power-scaled posterior of the inferred parameters of the second step, i.e. $p^\beta_{\rm ps}(\omega_R, \sigma_R \mid \cD_R)$. For the input parameter $\omega_R$, we observe that for $\beta = 0$ the posterior is equal to the prior. This is expected, since the likelihood of the simulation data is power-scaled by $\beta = 0$ and there is no information about $\omega_R$ in the real data. For $\beta = 0.5$, we observe the highest posterior density, with pdf values with up to 6. This is because we use the full likelihood of both data sources and thus the posterior is maximally informed. The highest posterior density at the true value $\omega_R^*$ is obtained for $\beta = 1$. In the right column, which shows the posterior of $\sigma_R$, we observe that as $\beta$ increases, the posterior density shifts to the right (i.e., higher $\sigma_R$ estimates). This means, that the estimated approximation error of the surrogate to the real data is lowest at $\beta = 0$ and highest at $\beta = 1$.

\subsection{Case Study 1.2: Modified train/test split}
Next, we demonstrate that the relative performance of posterior predictive weighting and power-scaling is scenario dependent. We slightly change the train/test split of the real-world data in case study 1, while keeping all other hyperparameters identical (see \secref{subsec:case_study_1}). Specifically, we use $N_R = 30$ evenly spaced input points $x_R \in [1,140]$ for real-world training data, while the out-of-distribution testing points are chosen from $x_R \in [140, 200]$. The other test scenarios, including out-of-sample (OOS) and combined setups are analogously constructed to case study 1.

\begin{figure*}[ht]
    \includegraphics[width=\textwidth]{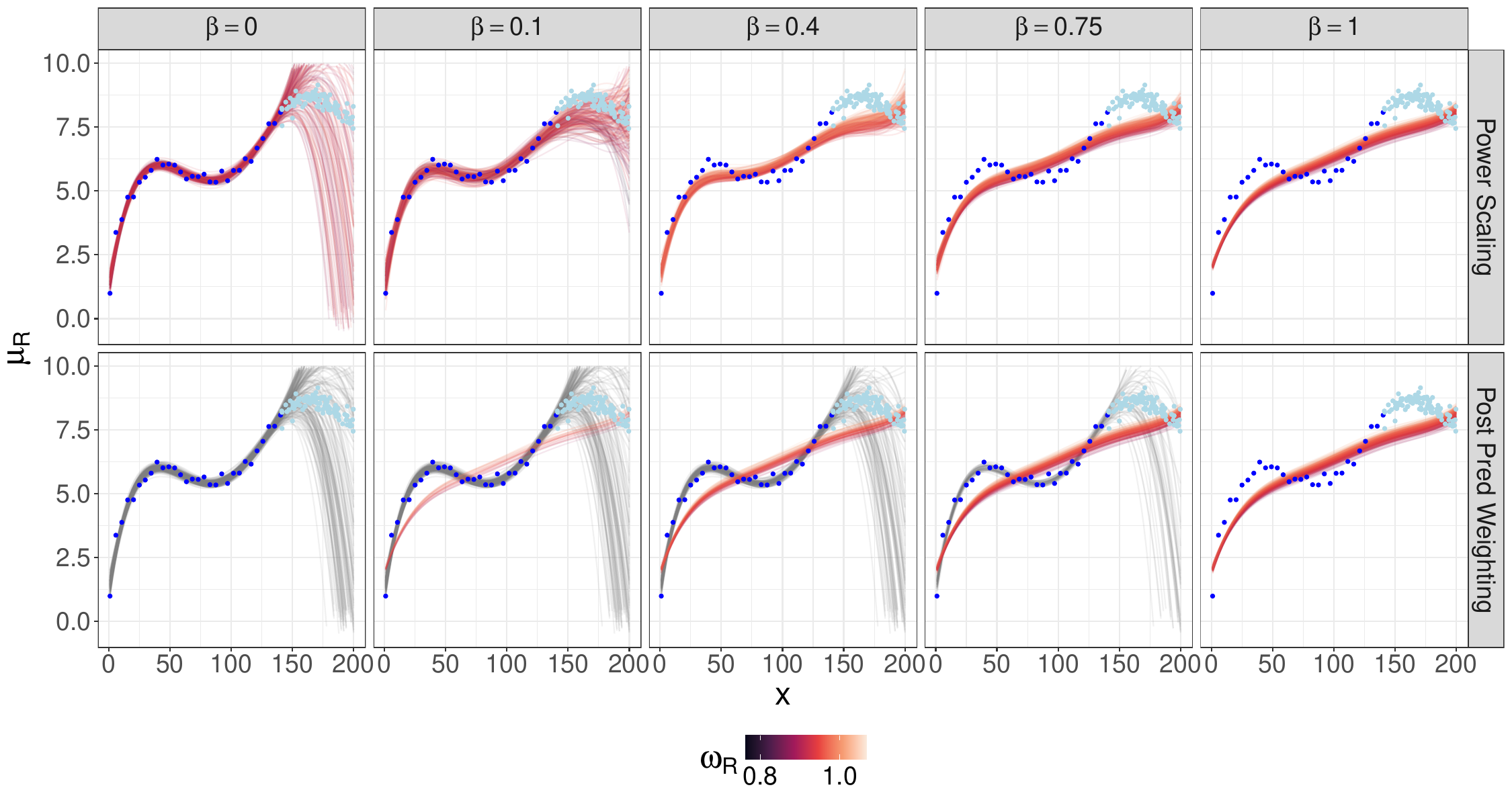}
    \caption{Case study 1.2: Predictive mean posteriors as obtained from the hybrid surrogates with varied weighting factor $\beta$. Top row: results using the power-scaling approach, bottom row: results from the posterior predictive weighting approach. Each line represents one of 200 posterior samples from the predictive mean posterior distribution, color-coded by the corresponding $\omega_R$ value. The dark and light blue points show train data and out-of-distribution test data, respectively.}
    \label{fig:case_study_1_2_posterior_epred}
\end{figure*}

\begin{figure*}[ht]
    \includegraphics[width=\textwidth]{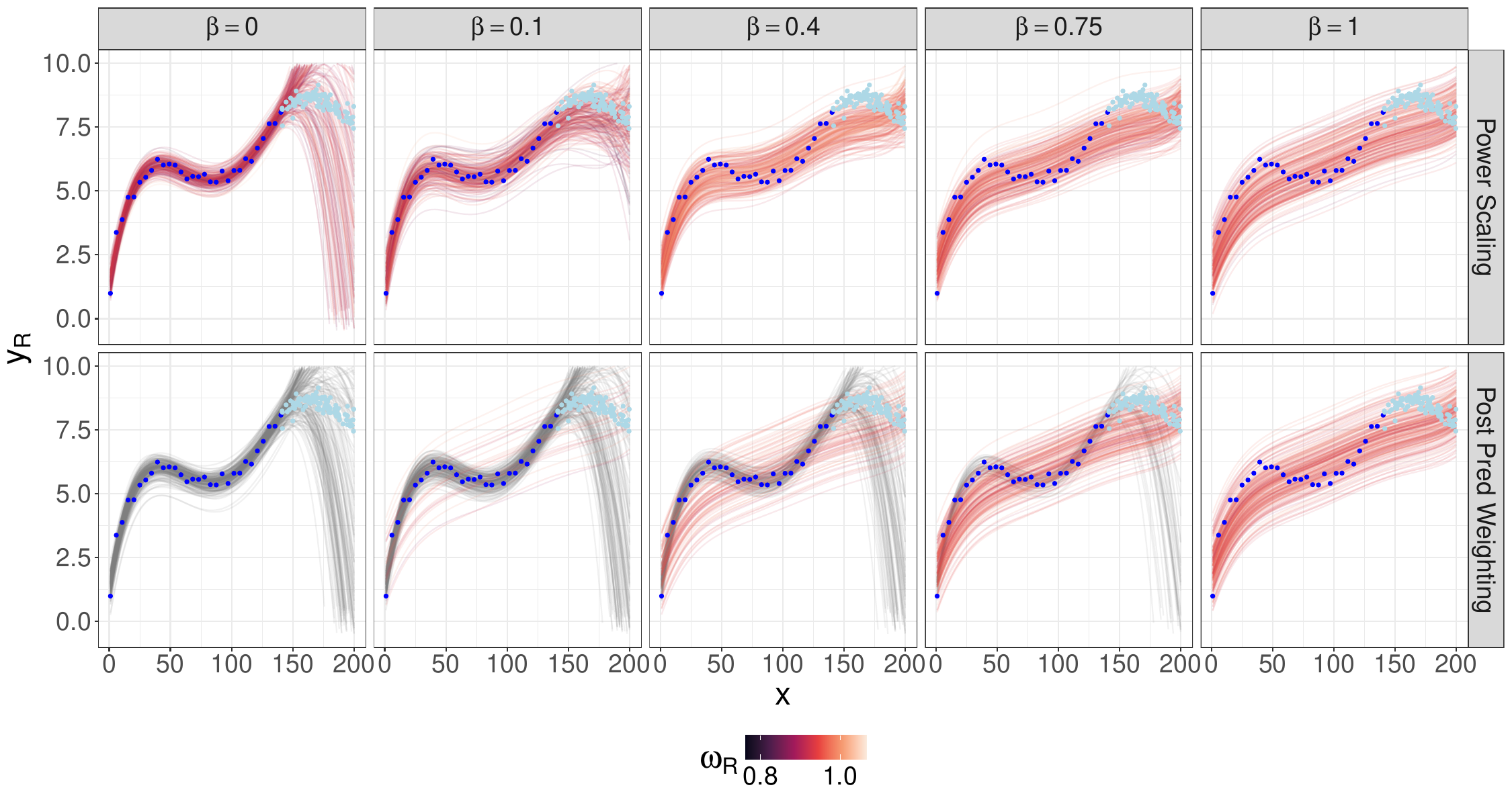}
    \caption{Case study 1.2: Posterior predictives as obtained from the hybrid surrogates with varied weighting factor $\beta$. Top row: results using the power-scaling approach, bottom row: results from the posterior predictive weighting approach. Each line represents one of 200 posterior samples from the predictive mean posterior distribution, color-coded by the corresponding $\omega_R$ value. The dark and light blue points show train data and out-of-distribution test data, respectively.}
    \label{fig:case_study_1_2_posterior_pred}
\end{figure*}

\begin{figure*}[ht]
    \includegraphics[width=\textwidth]{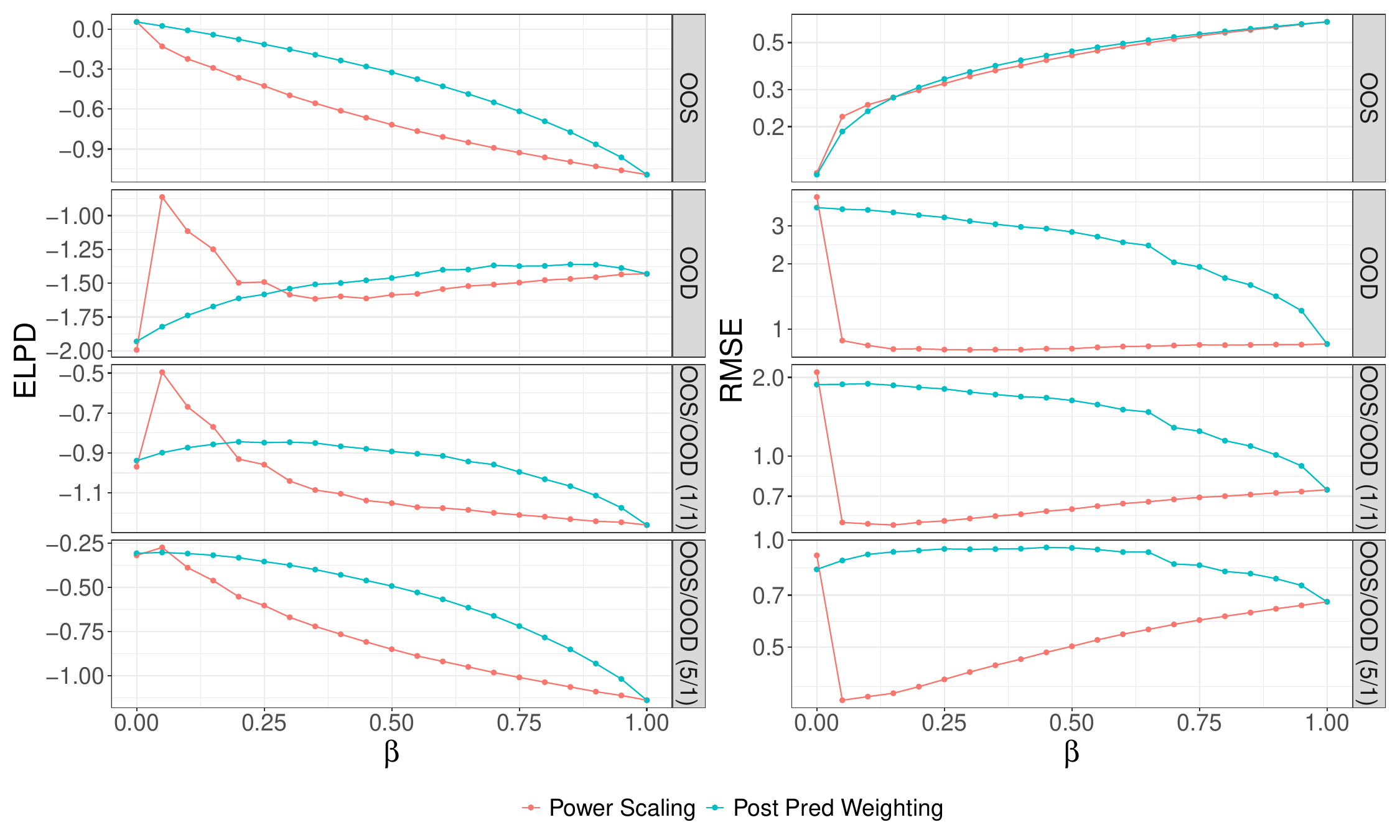}
    \caption{Case study 1.2: ELPD and RMSE of power-scaling and posterior predictive weighting on test data. The test data splits are out-of-distribution (OOD), out-of-sample in-distribution (OOS), and their combination (OOS/OOD) using 1/1 and 5/1 sample size ratios. For ELPD, higher is better; for RMSE, lower is better.}
    \label{fig:case_study_1_2_predictive_metrics}
\end{figure*}

In \figref{fig:case_study_1_2_posterior_epred}, we present the predictive mean posterior distributions, while \figref{fig:case_study_1_2_posterior_pred} shows the posterior predictive distributions. To evaluate predictive performance, we report the ELPD and RMSE metrics on the real test data in \figref{fig:case_study_1_2_predictive_metrics}. In this setup, the power-scaling approach achieves optimal performance in terms of ELPD, outperforming the posterior predictive weighting method. 


\subsection{Sensitivity to the relative sizes of simulation and real-world training data}
\begin{figure*}[ht]
\centering
\includegraphics[width=\textwidth]{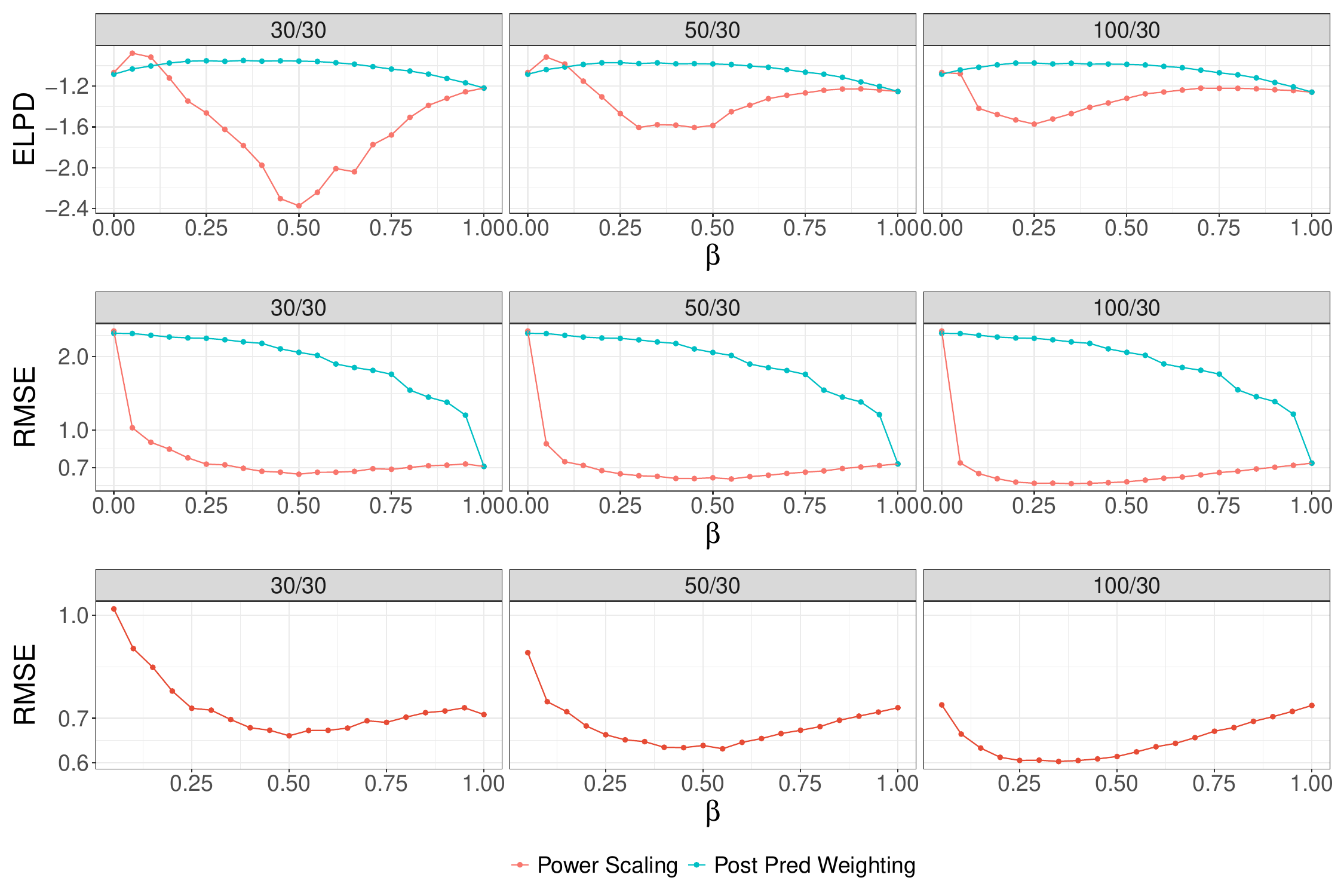}
\caption{
Case study 1: ELPD and RMSE of power-scaling and posterior predictive weighting across different simulation/real-data training ratios $(N_S / N_R)$.
The top row shows the expected log predictive density (ELPD), the middle row the root mean squared error (RMSE), and the bottom row a zoom-in of RMSE for the power-scaling approach.}
\label{fig:appendix_sim_real_ratio}
\end{figure*}

To further investigate the sensitivity of the proposed approaches to the relative sizes of the simulation and real-world training data, we varied the ratio $(N_S / N_R)$ across three settings: (30 / 30), (50 / 30), and (100 / 30) within the setup of case study 1.
\Cref{fig:appendix_sim_real_ratio} summarizes the results in terms of expected log predictive density (ELPD) and root mean squared error (RMSE) on the mixed OOS/OOD test scenario with 1/1 sample size ratio.

Across all training ratios, the posterior predictive weighting (PW) method yields comparable predictive performance over the full range of $\beta$ values. In terms of ELPD, both approaches show a characteristic pattern with a mild peak as $\beta$ increases from 0 toward intermediate values, indicating a modest benefit from blending simulation and real data. For the power-scaling (PS) approach, a slight decrease in performance is observed at certain intermediate $\beta$ values, reflecting the higher sensitivity of this method to the weighting configuration.

For RMSE, the PS method consistently achieves optimal performance for intermediate $\beta$ values, confirming that neither of the limiting cases $\beta = 0$ (purely data-driven surrogate) nor $\beta = 1$ (purely simulation-based surrogate) are optimal. The exact minimum shifts slightly across the three training-data ratios, but remains within the range $0.2 \leq \beta \leq 0.6$.
This illustrates an important advantage of the proposed approach: it enables the discovery of an optimal intermediate weighting that balances the contribution of both data sources -- something that would not have been possible without the proposed continuous weighting mechanisms.
\FloatBarrier

\subsection{Case Study 1 with GP Surrogate and Comparison to KOH}
\label{appendix:case_study_1_gp_koh}
 
To assess the generality of the proposed framework beyond polynomial surrogates, we replicate the setup of case study 1 (see \secref{subsec:case_study_1}) using a Gaussian Process (GP) as the surrogate model class. The GP surrogate employs a standard squared exponential kernel with automatic relevance determination \citep{neal1995bayesianlearning, rasmussen2008gaussianprocessesmachine}, parameterized by per-dimension length scales $\rho$ and marginal standard deviation $\alpha$, with priors $\rho \sim \text{Inv-Gamma}(5, 5)$ and $\alpha \sim \text{Normal}(0, 1)$.

In addition, we compare the power-scaling (PS) and posterior predictive weighting (PW) approaches against the Kennedy O'Hagan (KOH) calibration framework \citep{kennedyBayesianCalibrationComputer2001}, which represents the canonical GP-based approach for combining simulation and real-world data. The KOH model jointly estimates an emulator GP fitted to the simulation data and an additive discrepancy GP that captures the systematic deviation between the simulator and the real-world data-generating process. Unlike the proposed approaches, KOH does not expose a weighting factor $\beta$ to control the relative influence of the two data sources. Our KOH implementation extends the Stan code of \citet{chong2017bayesiancalibration} by replacing posterior predictive sampling with analytic GP conditioning, reducing computational cost at prediction time.
 
\paragraph{Posterior Predictive Plots}
 
\begin{figure*}[ht]
    \includegraphics[width=\textwidth]{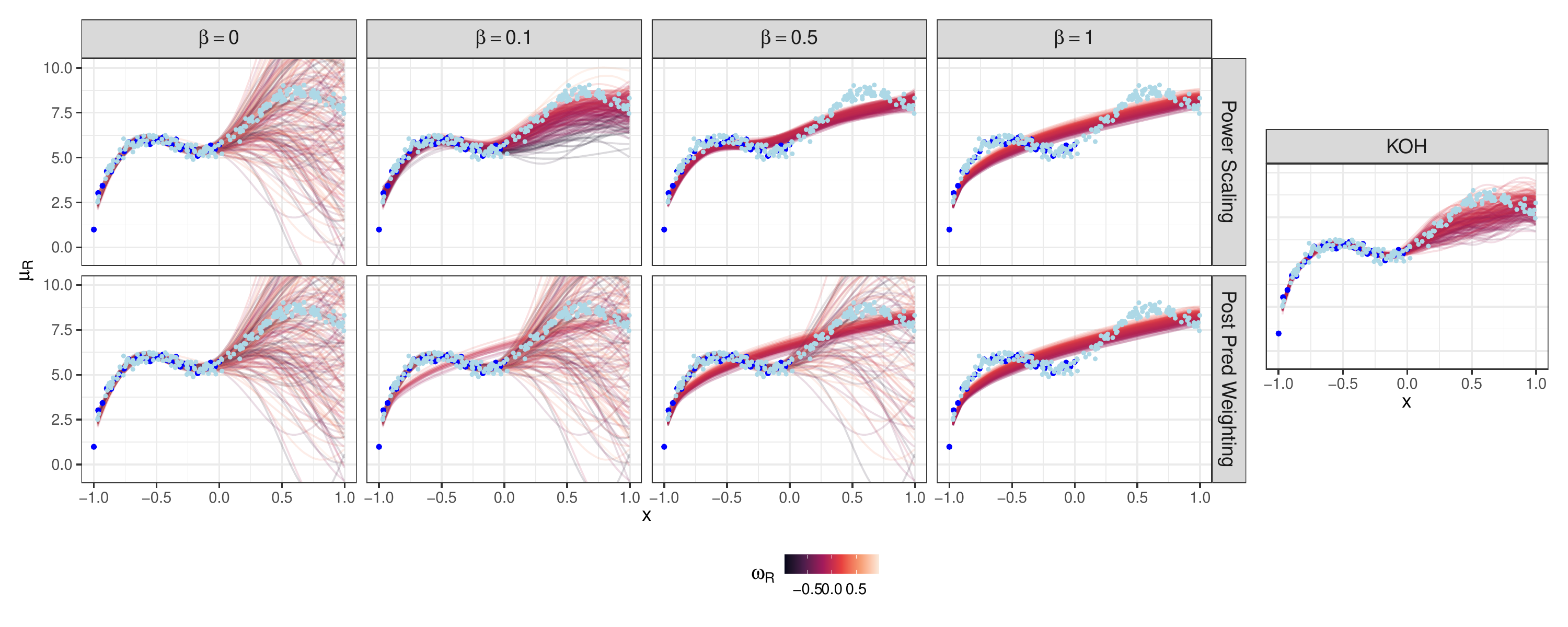}
    \caption{Case study 1 with GP surrogate: Predictive mean posteriors for the power-scaling (PS, top row) and posterior predictive weighting (PW, bottom row) approaches across $\beta \in \{0, 0.1, 0.5, 1\}$, with KOH shown as an additional panel on the right. Each line represents one of 200 posterior draws, color-coded by the corresponding $\omega_R$ value. Dark blue points indicate real-world training data; light blue points indicate out-of-distribution test data.}
    \label{fig:gp_posterior_pred_spaghetti}
\end{figure*}

\Cref{fig:gp_posterior_pred_spaghetti} displays 200 draws of the predictive mean posterior for the PS approach (top row) and the PW approach (bottom row) across four representative values of $\beta \in \{0, 0.1, 0.5, 1\}$, with the KOH prediction shown as an additional panel on the right. The overall qualitative behavior mirrors the PCE results: for $\beta = 0$ (data-driven GP), the in-distribution fit is close but the extrapolation into the OOD region is highly uncertain; for $\beta = 1$ (simulation-based GP), the global log-linear trend is well recovered but the periodic misspecification is not captured. At intermediate $\beta$ values, the PS surrogate mediates between the two extremes, progressively tightening the predictive bundle as $\beta$ increases. The KOH model, equipped with an explicit discrepancy GP, is able to fit both the training data and the smooth underlying trend, yielding a predictive distribution in the OOD region that is visually comparable to the PS approach at intermediate $\beta$ values. The PW approach (bottom row) retains the superposition character described in the main text: the relative weight of the simulation-based and data-driven components changes smoothly with $\beta$, making model discrepancies directly visible.
 
\begin{figure*}[ht]
    \includegraphics[width=\textwidth]{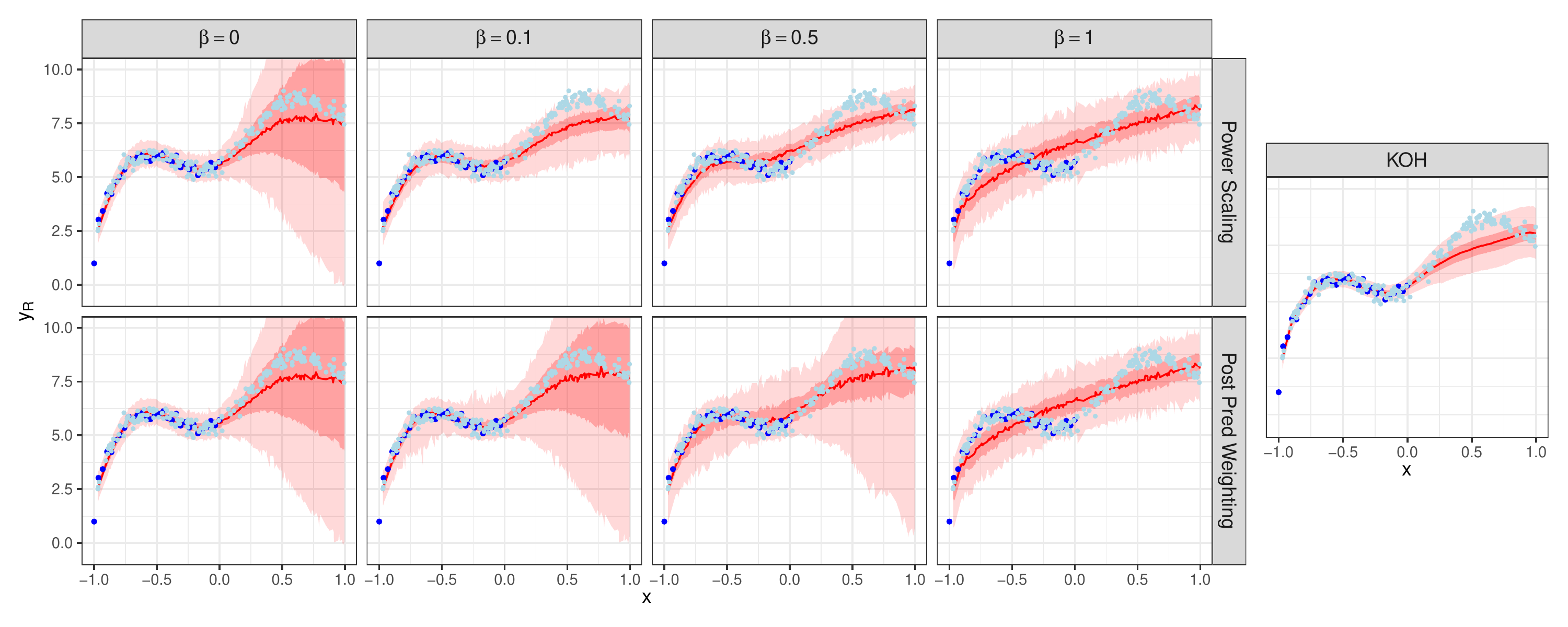}
    \caption{Case study 1 with GP surrogate: Posterior predictive distributions for PS (top row) and PW (bottom row) across $\beta \in \{0, 0.1, 0.5, 1\}$, with KOH on the right. Shaded bands show the 50\% and 95\% credible intervals; the solid line shows the median. Dark blue and light blue points correspond to training data and OOD test data, respectively.}
    \label{fig:gp_posterior_pred_ci}
\end{figure*}

\Cref{fig:gp_posterior_pred_ci} shows the corresponding posterior predictive distributions via the median and credible intervals. The PS approach tends to underestimate uncertainty for large $\beta$, consistent with the PCE findings in \secref{subsec:case_study_1}. The KOH model produces well-calibrated credible intervals in the OOD region, benefiting from the explicit discrepancy term. For both methods, the uncertainty in the OOD regime decreases as the contribution of simulation data increases, reflecting the structural regularization imposed by the simulation GP.
 
\paragraph{Predictive Performance}
 
\begin{figure*}[ht]
    \centering
    \includegraphics[width=0.9\textwidth]{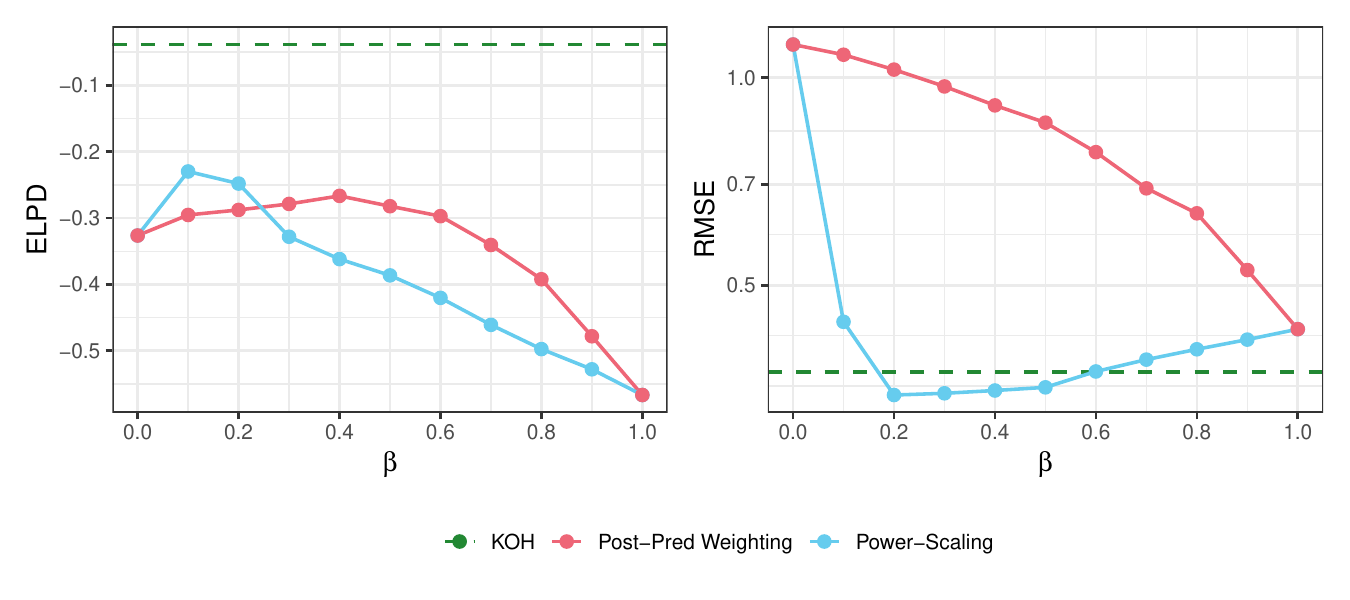}
    \caption{Case study 1 with GP surrogate: ELPD (left) and RMSE (right) on the OOS/OOD (1/1) test set as a function of $\beta$ for the power-scaling (PS, blue) and posterior predictive weighting (PW, pink) approaches. The dashed green line marks the performance of the KOH baseline. For ELPD, higher is better; for RMSE, lower is better.}
    \label{fig:gp_elpd_rmse}
\end{figure*}

\begin{figure}[ht]
    \centering
    \includegraphics[width=0.5\textwidth]{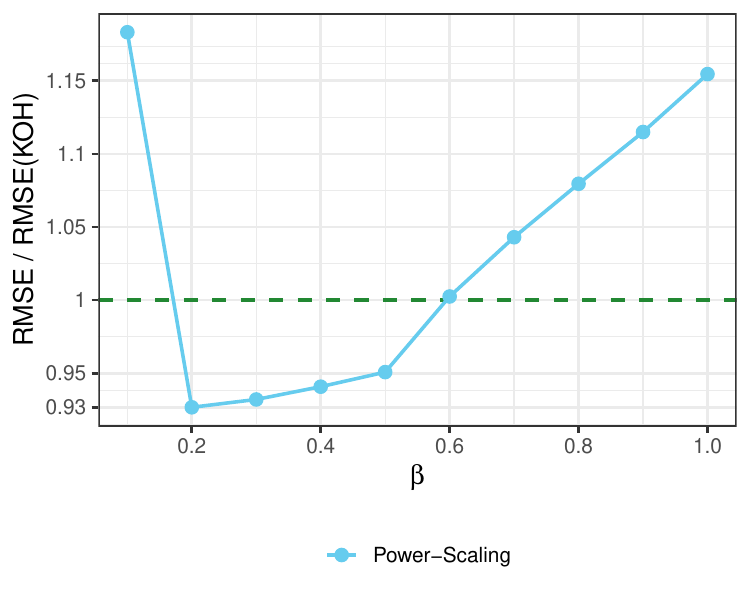}
    \caption{Case study 1 with GP surrogate: Ratio of power-scaling RMSE to KOH RMSE as a function of $\beta$. Values below the dashed green line (ratio $< 1$) indicate that the power-scaling approach outperforms KOH in terms of RMSE. The PS approach achieves a minimum ratio of approximately 0.93 at $\beta \approx 0.2$.}
    \label{fig:gp_rmse_ratio}
\end{figure}

\Cref{fig:gp_elpd_rmse} reports ELPD and RMSE on the OOS/OOD (1/1) test set for PS, PW, and KOH as a function of $\beta$. KOH (dashed green line) achieves the best ELPD among all evaluated approaches, indicating that the explicit discrepancy model provides a meaningful advantage for uncertainty quantification in this regime. In terms of RMSE, however, the PS approach attains values below the KOH reference for intermediate $\beta \in [0.2, 0.5]$. To highlight this finding, \figref{fig:gp_rmse_ratio} shows the ratio $\text{RMSE}_{\text{PS}}(\beta) / \text{RMSE}_{\text{KOH}}$ as a function of $\beta$. The PS approach outperforms KOH for $\beta \in [0.2, 0.5]$, achieving a minimum ratio of approximately 0.93 at $\beta = 0.2$, indicating that an appropriately tuned PS surrogate can recover the underlying mean function more accurately than the KOH discrepancy model in this setting.
 
Overall, these results confirm that our proposed framework is not restricted to PCE surrogates and transfers naturally to GP-based surrogates. At the same time, the comparison with KOH highlights the complementary strengths of the two paradigms: KOH excels at uncertainty quantification through its explicit discrepancy term, while the PS approach can achieve superior mean predictions, without requiring a separate discrepancy model.
 
\clearpage